\documentclass{article}

\usepackage{times}
\usepackage{graphicx}
\usepackage{calc}
\usepackage[notransparent]{svg}
\usepackage{amssymb}
\usepackage{amsmath}
\usepackage{xcolor}
\definecolor{cvprblue}{rgb}{0.21,0.49,0.74}
\usepackage[bookmarks=true]{hyperref}
\hypersetup{
	colorlinks=true,
	linkcolor=orange,
	filecolor=magenta,
	urlcolor=orange,
	citecolor=cvprblue,
}

\usepackage[utf8]{inputenc} 
\usepackage[T1]{fontenc}    
\usepackage{amsfonts}       
\usepackage{nicefrac}       
\usepackage{microtype}      

\usepackage{mathtools} 
\usepackage{cleveref}
\usepackage{booktabs} 
\usepackage{multirow} 
\usepackage{multicol} 
\usepackage[flushleft]{threeparttable}
\usepackage{colortbl} 
\usepackage{diagbox}  
\usepackage{siunitx}

\usepackage{enumitem} 
\usepackage{xspace} 
\usepackage[subrefformat=parens,labelformat=parens,font=small]{subcaption} 
\usepackage{wrapfig}
\usepackage{enumitem}

\usepackage{algorithm}
\usepackage{listings}
\usepackage{epigraph}
\usepackage{amsmath}

\crefname{table}{Tab.}{Tabs.}
\crefname{figure}{Fig.}{Figs.}
\crefname{section}{Sec.}{Secs.}
\crefname{appendix}{Appendix.}{Sections.}

\def\ie{\emph{i.e.}\xspace}

\newcommand{\ours}{Hume\xspace}

\newcounter{boldpara}
\newcommand{\boldparagraph}[1]{
    \refstepcounter{boldpara}
    \vspace{0.1em}\noindent{\bf #1}
}

\definecolor{mblue}{HTML}{367dbd}
\colorlet{colorFst}{mblue!45}     
\colorlet{colorSnd}{mblue!25}     
\colorlet{colorTrd}{yellow!30}    
\colorlet{colorLow}{darkgray!30}  
\definecolor{R1}{HTML}{E97451}
\definecolor{R2}{HTML}{008080}
\definecolor{R3}{HTML}{0047AB}
\colorlet{cmt}{darkgray!80}       
\colorlet{supp}{darkgray!50}      

\DeclareMathOperator*{\argmax}{arg\,max}

\definecolor{AC}{HTML}{E97451}
\definecolor{R1}{HTML}{008080}
\definecolor{R2}{HTML}{0047AB}
\definecolor{R3}{HTML}{EE82EE}
\definecolor{R4}{HTML}{6A5ACD}
\definecolor{Q}{HTML}{1f618d}

\let\titleold\title
\renewcommand{\title}[1]{\titleold{#1}\newcommand{\thetitle}{#1}}

\usepackage[preprint]{neurips_2025}

\title{\ours: Introducing System-2 Thinking in Visual-Language-Action Model}

\author{
    Haoming Song$^{1,2}$\thanks{Authors contributed equally: \href{mailto:haomingsong@sjtu.edu.cn}{haomingsong@sjtu.edu.cn}} \quad
    Delin Qu$^{3,2}$\footnotemark[1] \quad
    Yuanqi Yao$^{5,2}$ \quad
    Qizhi Chen$^{6,2}$ \quad
    Qi Lv$^2$ \quad
    Yiwen Tang$^{2}$ \quad \\
    \textbf{Modi Shi}$^4$ \quad
    \textbf{Guanghui Ren}$^4$ \quad
    \textbf{Maoqing Yao}$^4$ \quad
    \textbf{Bin Zhao}$^{2,7}$ \quad
    \textbf{Dong Wang}$^{2}$\footnotemark[2] \quad
    \textbf{Xuelong Li}$^{2,7}$\footnotemark[2] \\
    $^{1}$Shanghai Jiao Tong University \quad
    $^{2}$Shanghai AI Laboratory \quad
    $^{3}$Fudan University \quad
    $^{4}$AgiBot \quad \\
    $^{5}$INSAIT, Sofia University \quad
    $^{6}$Zhejiang University \quad
    $^{7}$Northwestern Polytechnical University \\
    \url{https://hume-vla.github.io} 
}

\begin{document}
\makeatletter
\let\@oldmaketitle\@maketitle
\renewcommand{\@maketitle}{\@oldmaketitle
  \begin{center}
  \vspace{-15pt}
  \captionsetup{type=figure}
  \setcounter{figure}{0}
  \includegraphics[width=1.0\textwidth]{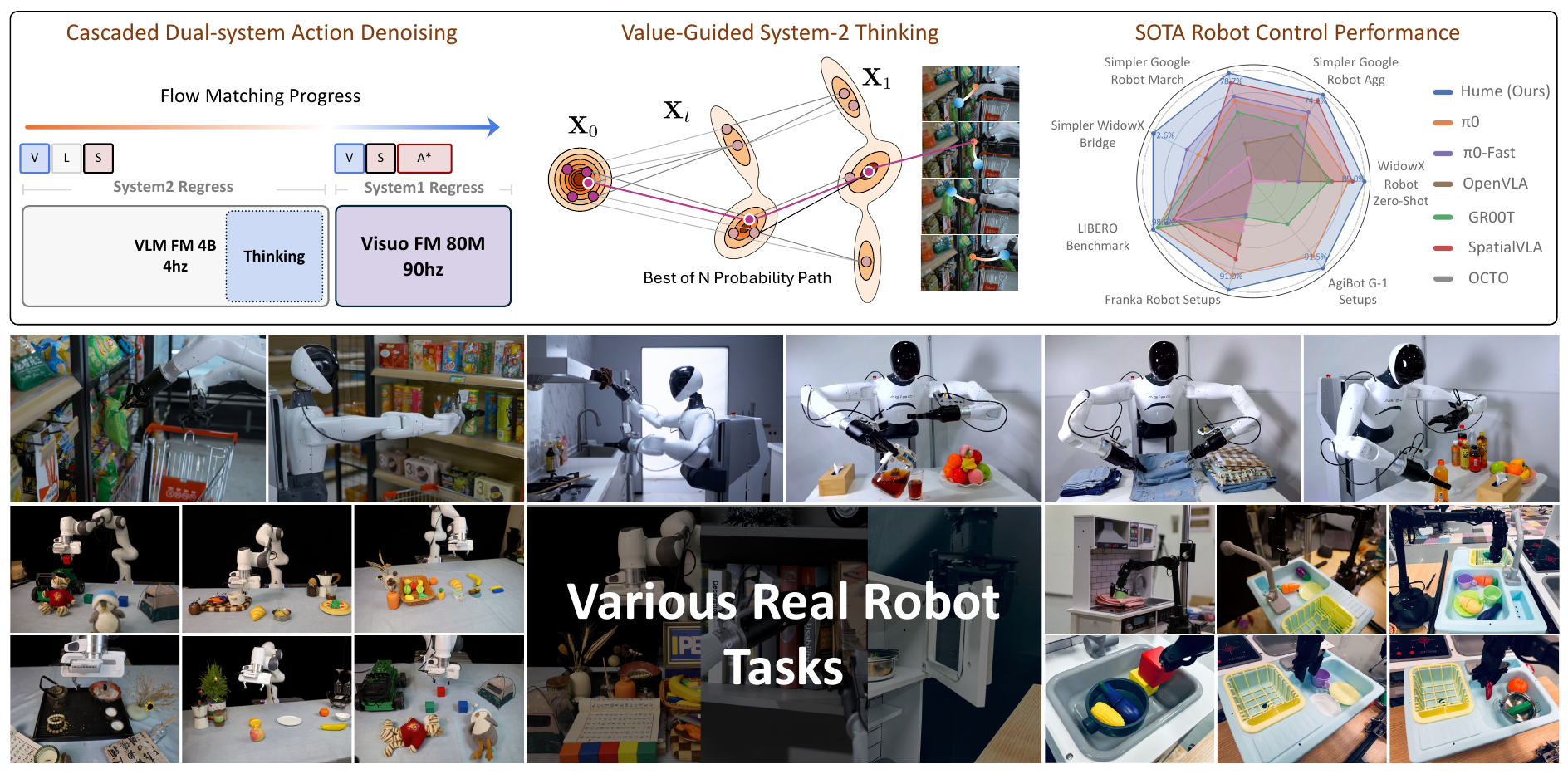}
    \caption{We present \ours, a dual-system vision-language-action model exploring human-like thinking capabilities for dexterous robot control. Equipped with value-guided System-2 thinking and cascaded action denoising, the model achieves superior complex reasoning and control capabilities. The model achieves state-of-the-art performance across a diverse range of evaluations and shows significantly advancement in complex robot control tasks.} 
    \vspace{-1ex}
    \label{fig:teaser}
  \end{center}
}
\makeatother

\maketitle
\footnotetext[2]{Corresponding author: \href{mailto:dongwang.dw93@gmail.com}{dongwang.dw93@gmail.com}}

\begin{abstract}
Humans practice slow thinking before performing actual actions when handling complex tasks in the physical world. 
This thinking paradigm, recently, has achieved remarkable advancement in boosting Large Language Models (LLMs) to solve complex tasks in digital domains.
However, the potential of slow thinking remains largely unexplored for robotic foundation models interacting with the physical world.
In this work, we propose \textbf{\ours}: a dual-system Vision-Language-Action (VLA) model with value-guided System-2 thinking and cascaded action denoising, exploring human-like thinking capabilities of Vision-Language-Action models for dexterous robot control.
System 2 of \ours implements value-guided thinking by extending a Vision-Language-Action Model backbone with a novel value-query head to estimate the state-action value of predicted actions. The value-guided thinking is conducted by repeat sampling multiple action candidates and selecting one according to state-action value.
System 1 of \ours is a lightweight reactive visuomotor policy that takes System 2 selected action and performs cascaded action denoising for dexterous robot control.
At deployment time, System 2 performs value-guided thinking at a low frequency while System 1 asynchronously receives the System 2 selected action candidate and predicts fluid actions in real time.
We show that \ours outperforms the existing state-of-the-art Vision-Language-Action models across multiple simulation benchmark and real-robot deployments. 
\end{abstract}
\section{Introduction}
\label{sec:intro}
\vspace{-4ex}
\setlength{\epigraphwidth}{0.45\textwidth}
\epigraph{\textit{A wise man proportions his belief to the evidence.}}{David Hume,\\\textit{An Enquiry Concerning Human Understanding}}
\vspace{-2ex}
Creating generalist robots to perform various tasks like humans in the physical world has been a long-standing goal~\citep{livision,brohan2023rt,liu2024fastumi, kim2024openvla, gao2025learning, pertsch2025fast, yao2024improving, zhang2025moma, gao2023orla, zhang2024decentralized,geminiroboticsteam2025geminiroboticsbringingai,black2024pi_0}.
Cognitive psychologists have revealed that humans conduct a deep, deliberate form of thinking when tackling complex problems, such as mathematical proofs or dinner making. This slow and reflective thought process is known as System-2 thinking, while the fast thinking process that relies on intuition is called System-1 thinking~\cite{kahneman2011thinking}.
Inspired by the dual process theory of human cognition, thinking and reasoning steps have been introduced to enhance LLMs' capability to solve complex problems in digital domains and achieved significant results.
Intuitively, generalist robots in the physical world also require similar System-2 slow thinking capabilities to perform dynamic, dexterous, and complex tasks~\citep{liu2024coherent,xia2024kinematic,wang2025more,wang2025skillnavenhancednavigationversatile}, while intuition-based fast System-1 thinking is not capable of tackling delicate, tenuous, and fallible robot action prediction~\citep{xia2025roboticpolicylearninghumanassisted,xia2025phoenix,yao2025think,Lv_2025_CVPR}. 
Therefore, a crucial question for generalist robot policies designed to solve complex robotic tasks in diverse scenarios is: \textbf{how to enable effective System-2 slow thinking in generalist robot policy for accurate action prediction?}

However, equipping a generalist robot policy with System-2 thinking capabilities poses two primary challenges.
First, thinking and reasoning techniques have mainly been demonstrated in text modality, while the delicate, tenuous robot actions lack of clear and consistent semantics, making it difficult to semantic Chain-of-Thought (CoT)~\cite{wei2022chain} thinking as in LLMs.
Second, the generalist robot policy needs to perform dexterous, complex tasks in real time. How to effectively balance the "slowness" of System-2 thinking against the "fastness" demanded by robot control is essential.
Recently, embodied chain-of-thought reasoning (ECoT)~\cite{Zawalski24-ecot} enable a VLA model to predict helpful intermediate reasoning text before choosing robot actions, improving the generalization and performance of VLA models. However, performing intermediate reasoning steps significantly slow down policy inference speed.
Moreover, dual-system architectures have been incorporated into Vision-Language-Action models in several works~\cite{han2024dual, zhang2024hirt, bu2024towards, wen2025dexvla,bjorck2025gr00t,shi2025hirobotopenendedinstruction}. Typical work Helix~\cite{helix} adopts a pretrained VLM backbone as System 2, while employing a smaller network as System 1 to outputs high-frequency actions for real time control. These approaches use either latent vectors or detailed language instructions as bridges to communicate between the two systems.
Despite their faster inference speed, these models' System 2 does not conduct effective thinking and reasoning to guide the System-1 action prediction.

In this work, we introduce a dual-system Vision-Language-Action model \ours, which powers the VLA model with System-2 thinking capabilities through value-guided repeat sampling and cascaded action denoising. 
The System 2 of \ours is built on top of a pre-trained vision-language model (VLM) and attach it with two specialized heads: a flow matching~\cite{lipman2022flow,lin2025beyond} denoising head to predict robot actions, and a value-query head to estimate the state-action value of predicted actions. It processes the robot’s observation and language instruction to predict long-horizon action chunk through action denoising head. Subsequently, the corresponding state-action value are estimated conditioning on predicted action chunk. The value-guided thinking is conducted by repeat sampling multiple action chunks and selecting one with the highest value.
For System 1 of \ours, it takes one short segment from the selected long-horizon action chunks from System 2, current visual observation, and robot states, then conducts cascaded action denoising to generate final fluid robot actions via a separate lightweight diffusion policy.
At deployment time, System 2 performs value-guided thinking at a low frequency (4 Hz) while System 1 asynchronously receives the System 2 selected action chunk and predicts fluid actions in real time (90 Hz).
Equipped with the proposed value-guided thinking and cascaded action denoising, \ours explores powerful System 2 slow thinking paradigm to enhance the VLA models.
We extensively evaluate and ablate \ours on both standard simulation benchmarks and real-robot platforms, including 21 real-world robot settings and 3 simulation environments.
To validate Hume's capability to solve complex robot control tasks with the assistance of System-2 Thinking, the test scenarios include variations in viewpoint, texture, lighting, layout, unseen objects, unseen environments, as well as the most challenging humanoid robot control tasks.
In summary, the main contributions of this work are three-folds:
\vspace{-2ex}
\begin{itemize}[noitemsep, topsep=10pt, leftmargin=10pt]
    \item We propose \ours, a dual-system generalist robot policy that explores System-2 slow thinking paradigm for Vision-Language-Action models.
    \item We introduce novel value-guided thinking and cascaded action denoising to seamlessly combine low frequency System 2 and high frequency System 1, resulting in effective thinking and reasoning in various robot deployments.
    \item \ours achieves state-of-the-art performance on multiple benchmarks and real-robot tests, achieving +4.4\% increase in success rate over $\pi_0$ on the LIBERO~\cite{liu2023libero} benchmark, +25.9\% in Simpler benchmark~\cite{li24simpler}, and +12.9\% improvement in real-world deployments.
\end{itemize}

\section{Related Work}
\label{sec:relate_work}
\vspace{-2ex}
\boldparagraph{Dual-System Vision-Language-Action Models.}Recently, several studies~\citep{livision,brohan2023rt,kim2024openvla,li2024cogact,zeng2025FSDrive,zheng2024tracevla,pmlr-v235-lv24a,li2024towards,zheng2025universal,liu2023learning,pertsch2025fast,bu2024towards,helix,han2024dual} have extended VLMs for robot control. RT-2~\citep{brohan2023rt} fine-tunes PaLI-X~\citep{chen2023pali} with discretized action tokens, while OpenVLA~\citep{kim2024openvla} adapts Prismatic VLM~\citep{karamcheti2024prismatic} on the OXE dataset~\citep{o2024open}. $\pi_0$~\citep{black2024pi_0} integrates PaliGemma with flow-matching for continuous actions. 
To address efficiency and integration challenges, dual-system architectures have emerged. 
HiRT~\citep{zhang2024hirt} runs VLMs at low frequencies while maintaining high-frequency vision-based control for real-time interaction. 
DexVLA~\citep{wen2025dexvla} uses diffusion action experts with embodiment curriculum learning across multiple robot types. 
GR00T N1~\citep{bjorck2025gr00t} features an end-to-end trained dual-system specifically for humanoid robots. 
Gemini Robotics~\citep{geminiroboticsteam2025geminiroboticsbringingai} builds on Gemini 2.0 with specialized models for control and reasoning.
HiRobot~\citep{shi2025hirobotopenendedinstruction} enables processing of complex instructions with situated feedback. These approaches improve efficiency, success rates, and adaptability compared to monolithic architectures.

\boldparagraph{System 2 and System-2 Thinking.}Numerous studies~\citep{wei2022chain,shinn2023reflexionlanguageagentsverbal,hao2023reasoninglanguagemodelplanning,shao2024deepseekmathpushinglimitsmathematical,zhang2025cross,NEURIPS2024_0d9dcd4e,miao2023selfcheck,wang2025morphmark,Tree_of_Thought} have explored System 2 reasoning approaches to enhance LLMs' problem-solving capabilities. Chain-of-Thought~\citep{wei2022chain,zhaxizhuoma2024alignbot} introduces intermediate reasoning steps before producing answers, while Tree-of-Thoughts~\citep{Tree_of_Thought} explores multiple solution paths with self-verification. Reflexion~\citep{shinn2023reflexionlanguageagentsverbal} enables verbal reflection on previous attempts. 
System-2 Thinking frameworks explicitly model human-like deliberative processes. SETS~\citep{chen2025setsleveragingselfverificationselfcorrection} combines self-critique with multiple reasoning paths and majority voting. 
SC-MCTS~\citep{gao2024interpretablecontrastivemontecarlo} combines multiple reward models for more robust tree search.
Addressing efficiency concerns, O1-Pruner~\citep{luo2025o1prunerlengthharmonizingfinetuningo1like} introduces length-penalty loss to create concise reasoning processes. While these approaches have improved reasoning in language tasks, their application to visual-language-action models remains largely unexplored.

\boldparagraph{Cascaded Denoising.}Cascaded Denoising~\cite{ho2021cascaded} was first proposed as a diffusion model~\cite{zhang2025revisiting} that generates higher-resolution images through a cascading approach. Starting with a diffusion model at low resolution, it continuously upsamples the generated images to obtain high-resolution results.
f-DM~\cite{gu2022fdm} applies inter-stage transformations for progressive signal restoration through function learning.
Cas-DM~\cite{an2024bring} cascades noise-prediction and image-prediction modules to integrate perceptual losses in diffusion models.
SCDM~\cite{chen2024spectral} cascades generation across spectral dimensions, reconstructing hyperspectral bands progressively.
HiFI~\cite{hur2025high} cascades consistent-resolution patches for memory-efficient high-resolution frame interpolation.
CDM-VTON~\cite{li2025cascaded} employs two-stage cascading for virtual try-on applications.
While these approaches have advanced capabilities in image domains, the application of cascaded denoising to integrate System 1 and System 2 remains largely unexplored.
\label{sec:method}
\section{Methodology}
\begin{figure*}[t]
    \begin{center}
        \includegraphics[trim=0.1ex 0 0 0, clip, width=1.0\textwidth]{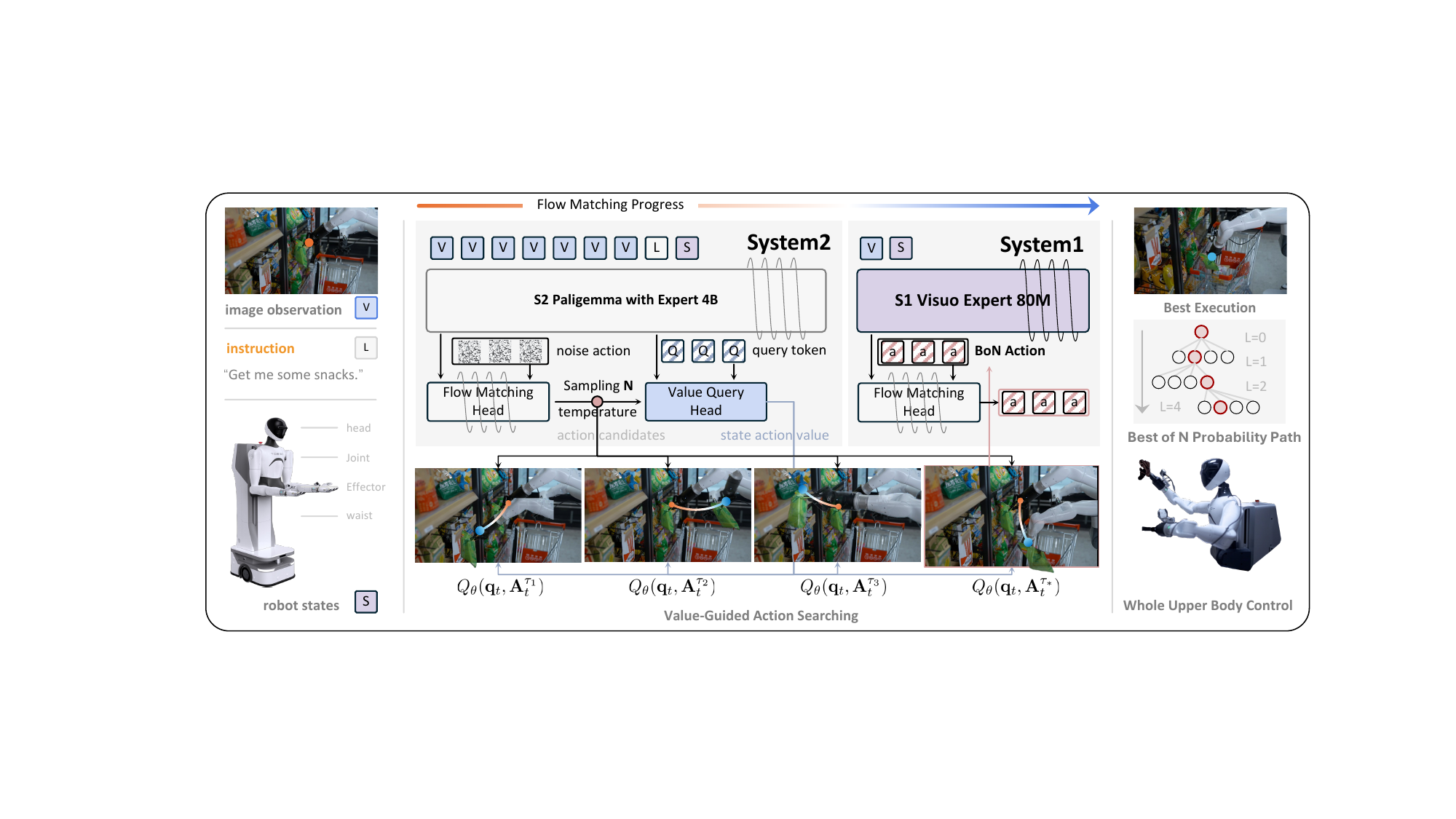}
    \end{center}
    \vspace{-1.5ex}
    \caption{\textbf{Overview of \ours.} \ours contains two systems working asynchronously. Given the observation, System 2 of \ours first generates $N$ candidate action chunks with different noise level, and the best-of-N candidate with the highest $Q$ value will be selected as the optimal candidate $\mathbf{A}_{t}^{\tau^*}$, which is segmented and conveyed to System 1 for continuous action denoising.}
    \label{fig:pipeline}
    \vspace{-2.5ex}
\end{figure*}
In this section, we describe \ours model architecture and its training and deployment strategy in detail.
The process of Value-Guided System-2 Thinking with the help of state-action value estimator was described in detail in ~\cref{sec:s2}. Next, we detail how the System 1 module and System 2 module cooperate asynchronously through the proposed cascaded action denoising in~\cref{sec:s1}.
Finally, the multi-stage training and deployment strategy of the model is explained in ~\cref{sec:training}.
\subsection{\textbf{Value-Guided System-2 Thinking}}
\label{sec:s2}
As shown in~\cref{fig:pipeline}, the System 2 module is instantiated as a vision-language-action model (VLA) built upon a pretrained Vision-Language Model.
Formally, the inputs of System 2 module consists of RGB images $\mathbf{i}_t = [\mathbf{I}^1_t, ... , \mathbf{I}^n_t]$ at time step $t$, natural language instructions $\mathbf{\ell_t}$, and robot state information $\mathbf{s}_t$. 
Similar as VLA models, we first augment the VLM backbone with an action denoising head to learn a mapping function $\mathcal{F}(\cdot)$ to generate candidate robot actions $\mathbf{A}_t$ from the observation $\mathbf{o}_t$, \ie, $\mathbf{A}_t=\mathcal{F}(\mathbf{o}_t)$.
Moreover, to empower \ours with System-2 Thinking ability, we attach the VLM backbone another value-query head, which is designed to estimate the state-action value $Q_{\theta}(\mathbf{q}_t, \mathbf{A}_t)$ of the candidate robot action $\mathbf{A}_t$.

\textbf{Candidate Actions Generation}
The candidate robot actions are generated by a action denoising head that aims to model the data distribution $p(\mathbf{A}_t|\mathbf{o}_t)$, where $\mathbf{o}_t$ consists of images $\mathbf{i}_t$, language instructions $\mathbf{\ell_t}$, and robot state information $\mathbf{s}_t$. It is implemented as a transformer-based flow matching denoising process that predicts the remaining action noise $\mathbf{v}_\theta(\mathbf{A}_t^\tau, \mathbf{o}_t)$ in the ``noisy action'' $\mathbf{A}_t^{\tau}$, where $\tau \in [0, 1]$ is the flow matching time step that represents the noise level of the action. 
Starting from a random noise $\mathbf{A}_{t}^{0} \sim \mathcal{N}(\mathbf{0}, \mathbf{I})$, the denoising head generates actions by gradually removing the noise from $\mathbf{A}_{t}^{0}$ to $\mathbf{A}_{t}^{1}$ step by step using the forward Euler method: $\mathbf{A}_{t}^{\tau+\delta} = \mathbf{A}_{t}^{\tau}+\delta \mathbf{v}_{\theta}\left(\mathbf{A}_{t}^{\tau}, \mathbf{o}_{t}\right)$, where $\delta$ is the size of the denoising step. 
In practice, we use 10 denoising steps, corresponding to $\delta=0.1$. 
During training, for the ground truth action $\mathbf{A}_t$ sampled from the dataset, the denoising head is optimized by minimizing the loss between the actual remaining noise $\epsilon - \mathbf{A}_t$ and the network output $\mathbf{v}_\theta(\mathbf{A}_t^\tau, \mathbf{o}_t)$ given the observation $\mathbf{o}_t$ and the noisy action $\mathbf{A}_t^\tau = \tau \mathbf{A}_t + (1 - \tau) \epsilon$ as input.

After training, conditioned on the same observation $\mathbf{o}_t$ at timestep $t$, the action denoising head generates $N$ candidate robot action chunks \
 $\mathbf{A}_{t}^{\tau_n} \in \
  \{
  \mathbf{A}_{t}^{\tau_1},\ 
  \mathbf{A}_{t}^{\tau_2},\ 
  \ldots,\
  \mathbf{A}_{t}^{\tau_N} 
\}$\
with different noise levels by integrating the learned vector field $\mathbf{v}_{\theta}\left(\mathbf{A}_{t}^{\tau}, \mathbf{o}_{t}\right)$ from $\tau = 0$ to $\tau = 1-(n-1)\xi$ separately:
\begin{align}
    \mathbf{A}_{t}^{\tau_n} & = \int_{0}^{1-(n-1)\xi} \mathbf{v}_{\theta}\left(\mathbf{A}_{t}^{\tau}, \mathbf{o}_{t}\right) d\tau + \mathbf{A}_{t}^{0},
\end{align}
where $\xi$ is used to control the noise gap between adjacent candidates, and $\mathbf{A}_{t}^{0}$ is the initial action sampled from the normal distribution $\mathbf{A}_{t}^{0} \sim \mathcal{N}(\mathbf{0}, \mathbf{I})$. 
Note that $n \in \{1,2,\ldots,N\}$, which leading to most of generated candidate actions  $\mathbf{A}_{t}^{\tau_n}$ is not fully denoised.

\textbf{State-Action Value Estimation}
The state-action values are estimated with the proposed value-query head built on the same VLM backbone via learning a latent conditioned Q function. The value-query head is composed of two critic networks estimating state-action values and one actor network for assisting the training of critic networks. 
Specifically, a special query token $\mathbf{q}_t$ is introduced and attached at the end of the VLM input sequence, which is a learnable token with the same embedding dimension as the language tokens.
Then, for one action chunk $\mathbf{A}_{t}$ (either ground-truth action $\mathbf{A}_{t}^{\text{GT}}$ or candidate actions $\mathbf{A}_{t}^{\tau_n}$), it is combined with this special query token $\mathbf{q}_t$ to feed into the value-query head.
Due to its last position of the input sequence, the query token $\mathbf{q}_t$ attends to all previous tokens and aggregates necessary information from the VLM inputs, \emph{i.e.}, current RGB images $\mathbf{i}_t = [\mathbf{I}^1_t, ... , \mathbf{I}^n_t]$ at time step $t$, natural language instructions $\mathbf{\ell_t}$. 
In this way, the value-query head estimate the state-action value $Q_{\theta}(\mathbf{q}_t, \mathbf{A}_t^{\tau_n})$ of the action chunk $\mathbf{A}_{t}^{\tau_n}$ conditioned on the input query token $\mathbf{q}_{t}$.
This value-query head is trained on pre-collected robot demonstration dataset with ground-truth action $\mathbf{A}_{t}$ via offline RL~\cite{levine2020offlinereinforcementlearningtutorial}. We construct the training dataset $\mathcal{D}$ using the reward function following~\cite{ptr}, where the rewards of last 3 transitions in one robot episode is defined as $+1$, and the rest is $0$. 
During training, we use the calibrated Q-learning algorithm~\cite{nakamoto2023calql} to optimize the value-query head. 

\begin{figure*}[t]
    \begin{center}
        \includegraphics[trim=0.5ex 0ex 0 0, clip, width=0.98\textwidth]{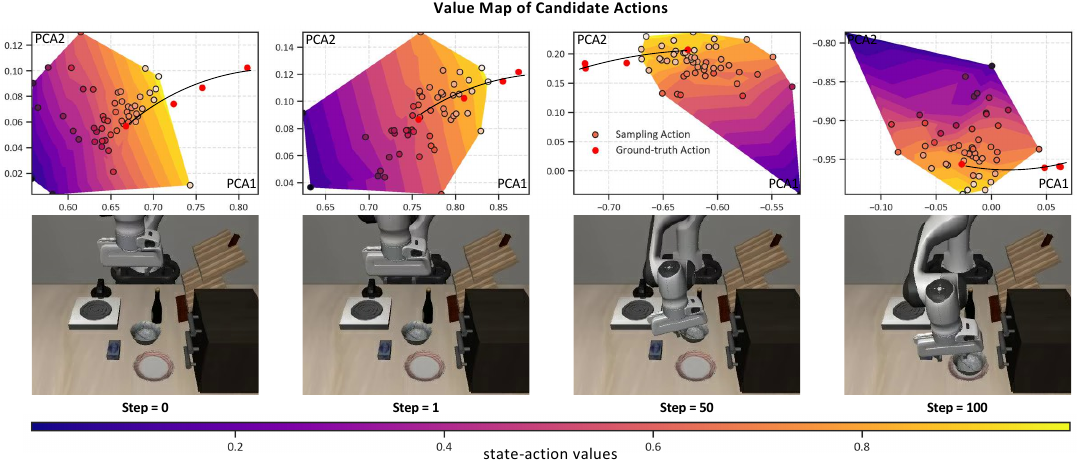}
    \end{center}
    \vspace{-1.5ex}
    \caption{\textbf{Value Map of Candidate Actions.} The candidate actions $\mathbf{A}_{t}^{\tau_n}$ sampled from System 2 and ground-truth actions $\mathbf{A}_{t}^{\text{GT}}$ are projected into the same two-dimensional space through Principal Component Analysis (PCA). The intensity of colors indicates the magnitude of state-action values $Q(\mathbf{q_t}, \mathbf{A}_{t}^{\tau_n})$ of candidate actions.}
    \label{fig:pca_map}
    \vspace{-2ex}
\end{figure*}

To verify our training pipeline can effectively optimize the value-query head, we project ground-truth actions $\mathbf{A}_{t}^{\text{GT}}$ and candidate actions $\mathbf{A}_{t}^{\tau_n}$ along with their corresponding state-action values $Q(\mathbf{q_t}, \mathbf{A}_{t}^{\tau_n})$ into the same space, generating the value map in~\cref{fig:pca_map}.
The value map consists of regions representing different magnitudes of state-action values, and the ground-truth actions are all located in the high-value region, which proves the value-query head can estimate reasonable state-action value.
The detailed analysis can be found in~\cref{sec:Value-Guided_Thinking_Visualization}

\textbf{Value-Guided Thinking}
The System-2 value-guided thinking is implemented with Best-of-N selection strategy and the selected action chunk is conveyed to System 1 for cascaded action denoising.
Specifically, conditioned on the same observation, the action denoising head generates $N$ candidate action chunks\
  $\{
  \mathbf{A}_{t}^{\tau_1},\ 
  \mathbf{A}_{t}^{\tau_2},\ 
  \ldots,\
  \mathbf{A}_{t}^{\tau_N}
\}$\
with different noise levels.
Then, these candidates are passed to the value-query head to estimate their state-action values.
Guided by the estimated state-action values, we select the action with the highest value as the optimal candidate $\mathbf{A}_{t}^{\tau^*}$ transferred to System 1, depicted as $\mathbf{A}_{t}^{\tau^*} = \mathop{\argmax}\nolimits_{\mathbf{A}_{t}^{\tau_n}} Q(\mathbf{q_t}, \mathbf{A}_{t}^{\tau_n})$, where $n=1 \ldots N$.
In~\cref{sec:Hume_Inference_Visualization}, we further visualize this process in the simulated Push-T task.

\subsection{\textbf{Cascaded Dual-system Action Denoising}}
\label{sec:s1}
In order to achieve rapid, reactive robot control, System 1 module needs to be lightweight and fast in inference. In detail, System 1 consists of a DINOv2-small visual encoder and a lightweight transformer for cascaded action denoising. Given the selected candidate action chunk $\mathbf{A}_{t}^{\tau^*}$ from System 2, the System 1 module takes the observation $\Tilde{\mathbf{o}}_{t+kh}$ (including current image $\mathbf{i}_t=\{\mathbf{I}_t^1,..,\mathbf{I}_t^n\}$, robot state $\mathbf{s}_t$), and sub-action chunks $\tilde{\mathbf{A}}_{t}$ segmented from the selected candidate $\mathbf{A}_{t}^{\tau^*}$ as input, and produces refined robot actions by continuously denoising on sub-action chunks $\tilde{\mathbf{A}}_{t}$.

Specifically, at timestep $t$, the selected action chunk from System 2 is $\mathbf{A}_{t}^{\tau^*} =[\mathbf{a}_t, \mathbf{a}_{t+1},...,\mathbf{a}_{t+H-1}]$, then $\mathbf{A}_{t}^{\tau^*}$ is segmented into $K:= H/h$ sub-action chunks $\{\Tilde{\mathbf{A}}_t, \Tilde{\mathbf{A}}_{t+h}, \ldots, \Tilde{\mathbf{A}}_{t+(K-1)h}\}$ with a horizon of $h$.
The System 1 sequentially performs cascaded denoising on these sub-action chunks with observation $\Tilde{\mathbf{o}}_{t+kh}$. Note that System 1 is much faster than System 2, so that System 1 could finish cascaded denoising on all sub-action chunks before next action chunk $\mathbf{A}_{t}^{\tau^*}$ arriving.
System 1 module is trained with the same flow matching loss using the action denoising head of System 2:
\begin{align}
    \label{eq:flow_matching_s1}
    L^\omega(\theta) &= \mathbb{E}_{p(\Tilde{\mathbf{A}}_{t+kh} | \Tilde{\mathbf{o}}_{t+kh}), q(\Tilde{\mathbf{A}}_{t+kh}^\omega | \Tilde{\mathbf{A}}_{t+kh})} || \mathbf{v}_\theta(\Tilde{\mathbf{A}}_{t+kh}^\omega, \Tilde{\mathbf{o}}_{t+kh}) - \mathbf{u}(\Tilde{\mathbf{A}}_{t+kh}^\omega | \Tilde{\mathbf{A}}_{t+kh}) ||^2,
\end{align}
while the superscript $\omega$ represents the flow matching timestep in System 1.

Note that, the generated candidate action chunks from System 2 are not fully denoised, \emph{i.e.}, $\Tilde{\mathbf{A}}_{t+kh}^{\tau^*}\ne\Tilde{\mathbf{A}}_{t+kh}^{1}$, requiring the continuous denoising for accurate action prediction.
Following the continuous denoising~\cite{ho2021cascaded} from image generation, System 1 refine the action by intergrating the learned vector field from $\omega = 0$ to $\omega = 1$. Instead of starting with random noise, the integration process of System 1 starting with the sub-action chunk $\Tilde{\mathbf{A}}_{t+kh}^{\tau^*}$:
\begin{align}
    \Tilde{\mathbf{A}}_{t+kh}^{\omega} & = \int_{0}^{\omega} \mathbf{v}_{\theta}\left(\Tilde{\mathbf{A}}_{t+kh}^{\omega}, \Tilde{\mathbf{o}}_{t+kh}\right) d\omega + \Tilde{\mathbf{A}}_{t+kh}^{\tau^*},
\end{align}
where $\mathbf{v}_{\theta}\left(\Tilde{\mathbf{A}}_{t+kh}^{\omega}, \mathbf{o}_{t+kh}\right)$ is the vector field learned by System 1. Using the forward Euler method $\Tilde{\mathbf{A}}_{t+kh}^{\omega+\sigma} = \Tilde{\mathbf{A}}_{t+kh}^{\omega}+\delta \mathbf{v}_{\theta}\left(\Tilde{\mathbf{A}}_{t+kh}^{\omega}, \Tilde{\mathbf{o}}_{t+kh}\right)$,
System 1 produces the final denoised action $\Tilde{\mathbf{A}}_{t+kh}^{1}$ with 10 denoise steps (corresponding to $\delta=0.1$).
After all K sub-action chunks have been processed by System 1, System 2 will generate a new selected action chunk ${\mathbf{A}}_{t+H}^{\tau^*}$, and System 1 will continue to refine segments from the new selected action chunk.

\subsection{\textbf{Training and Deployment Strategy}}
The training process of \ours contains two stages. In the first stage, the VLM backbone and the action denoising head of System 2 are trained first using the flow matching loss similar to~\cref{eq:flow_matching_s1} to ensure System 2 can predict reliable actions.
In the second stage of training, the VLM backbone and the action denoising head of System 2 are frozen, while System 1 and value-query head of System 2 are trained from scratch. 
The training objective of the Value-Query Head is to minimize the Bellman error with a regularization term, which is defined as:
\begin{align}
    \label{eq:cql_training}
    \!\!\!\!\!\min_\theta {\alpha\mathcal{R}(\theta)} + \frac{1}{2} {\mathbb{E}_{\mathbf{q}_t, \mathbf{A}_t, \mathbf{q}_t^\prime\sim \mathcal{D}}\left[\left(Q_\theta(\mathbf{q}_t, \mathbf{A}_t) - \mathcal{B}^\pi\bar{Q}(\mathbf{q}_t, \mathbf{A}_t)\right)^2 \right]},
\end{align}
where $\mathcal{R}(\theta)$ is a calibrated conservative regularizer that aims to prevent overestimation in the Q-values, $\mathcal{R}(\theta) := \mathbb{E}_{\mathbf{q}_t \sim \mathcal{D}, a \sim \pi}\left[\max \left(Q_{\theta}(\mathbf{q}_t, \mathbf{A}_t), Q^{\mu}(\mathbf{q}_t, \mathbf{A}_t)\right)\right]-\mathbb{E}_{\mathbf{q}_t, \mathbf{A}_t \sim \mathcal{D}}\left[Q_{\theta}(\mathbf{q}_t, \mathbf{A}_t)\right]$, and $\mathcal{B}^\pi \bar{Q} (\mathbf{q}_t, \mathbf{A}_t)$ is the backup operator applied to the delayed target Q-network $Q_{\bar{\theta}}$.

During the inference phase, the System 2 and System 1 modules are cooperating at asynchronous mechanism to boost up the overall control frequency. 
Specifically, at the initial timestep $t$, the action denoising head of System 2 generates $N=5$ multiple action chunks\
$\mathbf{A}_{t}^{\tau_n} \in \
\{
\mathbf{A}_{t}^{\tau_1},\ 
\mathbf{A}_{t}^{\tau_2},\ 
\ldots,\
\mathbf{A}_{t}^{\tau_N} 
\}$\
with horizon of $H=30$ as candidates at 4 Hz. 
Then the selected optimal action candidate $\mathbf{A}_{t}^{\tau^*}$ is stored in a shared queue.
Then $\Tilde{\mathbf{A}}_{t}^{\tau^*}$, a sub-action chunk with horizon $h=15$, is segmented from the first $h$ steps of $\mathbf{A}_{t}^{\tau^*}$ and passed into System 1. 
System 1 removes the remaining noise from $\Tilde{\mathbf{A}}_{t}^{\tau^*}$, and produces the fully denoised action $\Tilde{\mathbf{A}}_{t}$ at 6 Hz and execute all $h=15$ actions on real robot immediately, resulting in a overall 90 Hz robot action control frequency.
After the robot executes all $K=H/h=2$ sub-action chunks in $\Tilde{\mathbf{A}}_{t}$, System 1 repeatly get the newest selected action chunks from the shared queue for subsequent action denoising.
Due the different working frequencies of System 2 and System 1, they asynchronously cooperate to achieve a balance between slow, human-like thinking and fast, reactive real robot control.
\label{sec:training}
\section{Experiment}
\label{sec:experiment}
\begin{figure*}[h]
    \vspace{-1ex}
    \begin{center}
       \includegraphics[width=\linewidth]{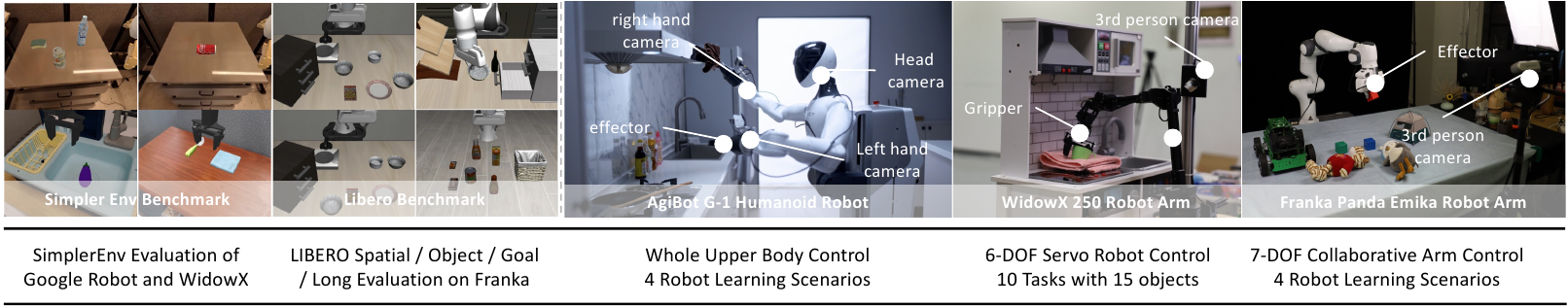}
   \end{center}
  \caption{\textbf{Experiments setup on WidowX, AgiBot G-1 and Franka Robot.} We evaluate \ours across 3 simulation
environments and 3 different real-world robotic platforms, covering 15 robot learning scenarios and 21 real-world manipulation tasks.}
   \label{fig:whole_exp_setup}
\end{figure*}

Our experiments aim to evaluate whether \ours, as a dual-system Vision Language Action Model, can effectively utilize System-2 Thinking to solve complex robot control tasks.
Our extensive experiments include evaluating the model's ability to perform complex manipulation tasks on various robotic platforms in both simulated and real-world environments, including humanoid robots.
\ours is compared with previous state-of-the-art generalist policies and alternative designs of various model components.
Specifically, our experiments aim to answer the following research questions:
\begin{enumerate}[noitemsep, topsep=0pt, leftmargin=10pt]
    \item How is \ours's capability to learn multiple tasks on standard simulation benchmarks? 
    \item Can \ours effectively solve a variety of complex robot control tasks in the real world?
    \item To what extent do value-guided thinking and cascaded denoising improve the performance?
\end{enumerate}

To answer these questions, as shown in~\cref{fig:whole_exp_setup}, we tested \ours's capabilities across a diverse range of representative robot learning scenarios, including 
3 simulation environments and 3 different real-world robotic platforms, covering 15 robot learning scenarios and 21 real-world manipulation tasks.
First, we evaluated \ours's capability to finish multiple tasks in SimplerEnv~\citep{li24simpler} and LIBERO~\cite{liu2023libero} simulation benchmarks, validating that \ours's design can effectively accomplish multiple tasks in simulated environments.
Second, we extensively tested \ours's capability to control 3 real-world robotic platforms, WidowX, Franka, and AgiBot G-1, in completing tasks of varying difficulty, effectively validating \ours's generalization capability in aspects such as object positions, language descriptions, deformable objects, and long-horizon operations in the real world. Finally, we conducted comprehensive ablation experiments on the model design in both simulation and real-world environments to validate the design choices in \ours.

\subsection{\textbf{Multitask Evaluation on Simulation Benchmarks}} 
\label{sec:simulation}
\textbf{Evaluation Setups and Baselines.} To assess the robustness of \ours in diverse environmental variations, we employ the SimplerEnv~\citep{li24simpler} simulation benchmark to evaluate visual matching and variant aggregation metrics. SimplerEnv features WidowX and Google Robot setups, providing diverse manipulation scenarios with varied lighting, color, textures, and robot camera pose conditions, bridging the visual appearance gap between real and simulated environments. We compare our model with the latest state-of-the-art generalist manipulation policies, including RT-1~\citep{brohan2022rt}, RT-1-X~\cite{o2024open}, RT-2-X~\cite{o2024open}, Octo~\cite{team2024octo}, OpenVLA~\cite{kim2024openvla}, HPT~\citep{wang2024scaling}, TraceVLA~\cite{zheng2024tracevla}, RoboVLM~\cite{li2023generalist}, SpatialVLA~\cite{qu2025spatialvlaexploringspatialrepresentations}, GR00T~\cite{bjorck2025gr00t}, $\pi_0$-FAST~\cite{pertsch2025fast}, and $\pi_{0}$~\cite{black2024pi_0}. 

\noindent \textbf{Evaluation Results.} \cref{tab:libero} present the LIBERO~\citep{liu2023libero} experimental results. We observe that \ours can be effectively adapted to tasks in the LIBERO environments, as it obtains the highest average success rate of 98.6\% and the first rank across all the policies. 
In particular, \ours achieves a remarkable 96.7\% success rate (+11.5\% over $\pi_0$, +6.1\% over GR00T) on the LIBERO-Long task, which consists of long-horizon tasks, demonstrating the model's strong long-term planning capabilities. 
\cref{tab:simplerenv_combined} presents the SimplerEnv experimental results on WidowX and Google robot tasks. \ours also achieves state-of-the-art performance on WidowX multitasks, with an average success rate of 72.6\%, representing a significant improvement compared to all current generalist manipulation policies (+32.5\% over $\pi_0$, +64.8\% over OpenVLA).
Similarly, \ours achieves an average success rate of 76.4\% (+19.6\% over $\pi_0$) on Google robot tasks.
In summary, \ours demonstrates its versatility as a generalist robot control policy, achieving better performance across various tasks.

\begin{table*}
    [h]
    \centering
    \caption{\textbf{LIBERO Benchmark Results.} We present the success
    rate (SR) and standard error for each method across four task suites, which
    are averaged over three random seeds with 500 trials. \ours achieve the highest average success rate and ranking, followed by OpenVLA-OFT and $\pi_{0}$.}
    \label{tab:libero} \resizebox{\textwidth}{!}{\begin{tabular}{l|cc|cc|cc|cc|cc}
        \toprule                                                      & \multicolumn{2}{c|}{LIBERO-Spatial} & \multicolumn{2}{c|}{LIBERO-Object} & \multicolumn{2}{c|}{LIBERO-Goal} & \multicolumn{2}{c|}{LIBERO-Long} & \multicolumn{2}{c}{Average} \\
        \cline{2-3} \cline{4-5} \cline{6-7} \cline{8-9} \cline{10-11} & SR ($\uparrow$)                     & Rank ($\downarrow$)                & SR ($\uparrow$)                  & Rank ($\downarrow$)              & SR ($\uparrow$)            & Rank ($\downarrow$) & SR ($\uparrow$)           & Rank ($\downarrow$) & SR ($\uparrow$)           & Rank ($\downarrow$) \\
        \midrule Diffusion Policy~\citep{chi2024diffusionpolicy}                                    & 78.5 $\pm$ 1.1\%                    & 6                                  & 87.5$\pm$ 0.7\%                  & 6                                & 73.5$\pm$ 1.2\%            & 6                   & 64.8$\pm$ 1.3\%           & 5                   & 76.1 $\pm$ 0.7\%          & 6                   \\
        OpenVLA-OFT~\cite{kim2025fine}                                                 & 97.6$\pm$ 0.9\%                     & 2                                  & 98.4$\pm$ 0.8\%                  & 3                                & 97.9 $\pm$ 1.0\%           & 2                   & 94.5$\pm$ 1.3\%           & 2                   & 97.1$\pm$ 0.6\%           & 2                   \\
        $\pi_{0}$~\cite{black2024pi_0}                                & 96.8 $\pm$ 0.8\%                    & 3                                  & 98.8 $\pm$ 0.9\%                 & 2                                & 95.8 $\pm$ 1.1\%           & 3                   & 85.2 $\pm$ 1.2\%          & 4                   & 94.2 $\pm$ 0.9\%          & 3                   \\
        $\pi_{0}$-FAST~\cite{pertsch2025fast}                         & 96.4 $\pm$ 0.7\%                    & 4                                  & 96.8 $\pm$ 0.7\%                 & 5                                & 88.6 $\pm$ 1.0\%           & 5                   & 60.2 $\pm$ 1.4\%          & 6                   & 85.5 $\pm$ 1.0\%          & 4                   \\
        GR00T N1~\cite{bjorck2025gr00t}                               & 94.4 $\pm$ 0.9\%                    & 5                                  & 97.6 $\pm$ 1.0\%                 & 4                                & 93.0$\pm$ 1.2\%            & 4                   & 90.6$\pm$ 1.0\%           & 3                   & 93.9 $\pm$ 1.1\%          & 5                   \\
        \rowcolor[HTML]{EFEFEF}\ours                                  & \textbf{98.6 $\pm$ 0.2\%}           & 1                                  & \textbf{99.8 $\pm$ 0.1\%}        & 1                                & \textbf{99.4 $\pm$ 0.3\%}  & 1                   & \textbf{96.7 $\pm$ 0.9}\% & 1                   & \textbf{98.6 $\pm$ 0.7\%} & 1                   \\
        \bottomrule
    \end{tabular}}
\end{table*}
\begin{table*}[t]
    \centering
    \caption{\textbf{Evaluation results across different policies on SimplerEnv}. We evaluate \ours on 4 tasks of WidowX and 3 tasks on Google-EDR in SimplerEnv}
    \vspace{-1ex}
    \resizebox{\textwidth}{!}{
    \begin{tabular}{l|cccccccccc}
        \toprule
        \multicolumn{11}{c}{\textbf{SimplerEnv on WidowX Robot Tasks}} \\
        \midrule
        \multirow{2}{*}{Model} & \multicolumn{2}{c}{\textbf{Put Spoon on Towel}} & \multicolumn{2}{c}{\textbf{Put Carrot on Plate}} & \multicolumn{2}{c}{\textbf{Stack Green Block on Yellow Block}} & \multicolumn{2}{c}{\textbf{Put Eggplant in Yellow Basket}} & \textbf{\#Overall} \\
        & \begin{tabular}[c]{@{}c@{}}Grasp Spoon\end{tabular} & \begin{tabular}[c]{@{}c@{}}Success\end{tabular} & \begin{tabular}[c]{@{}c@{}}Grasp Carrot\end{tabular} & \begin{tabular}[c]{@{}c@{}}Success\end{tabular} & \begin{tabular}[c]{@{}c@{}}Grasp Green Block\end{tabular} & \begin{tabular}[c]{@{}c@{}}Success\end{tabular} & \begin{tabular}[c]{@{}c@{}}Grasp Eggplant\end{tabular} & \begin{tabular}[c]{@{}c@{}}Success\end{tabular} & Average \\
        \cmidrule{1-10}
        RT-1-X~\cite{o2024open} & 16.7\% & 0\% & 20.8\% & 4.2\% & 8.3\% & 0\% & 0.0\% & 0\% & 6.3\% \\
        Octo-Base~\cite{team2024octo} & 34.7\% & 12.5\% & 52.8\% & 8.3\% & 31.9\% & 0\% & 66.7\% & 43.1\% & 31.3\% \\
        Octo-Small~\cite{team2024octo} & \textbf{77.8\%} & 47.2\% & 27.8\% & 9.7\% & 40.3\% & 4.2\% & 87.5\% & 56.9\% & 43.9\% \\
        OpenVLA~\cite{kim2024openvla} & 4.1\% & 0\% & 33.3\% & 0\% & 12.5\% & 0\% & 8.3\% & 4.1\% & 7.8\% \\
        RoboVLM~\cite{li2023generalist} & 54.2\% & 29.2\% & 25.0\% & 25.0\% & 45.8\% & 12.5\% & 58.3\% & 58.3\% & 38.5\% \\
        SpatialVLA~\cite{qu2025spatialvlaexploringspatialrepresentations} & 25.0\% & 20.8\% & 41.7\% & 20.8\% & 58.3\% & 25.0\% & 79.2\% & 70.8\% & 42.7\% \\
        $\pi_{0}$~\cite{black2024pi_0} & 45.8\% & 29.1\% & 25.0\% & 0\% & 50.0\% & 16.6\% & 91.6\% & 62.5\% & 40.1\% \\
        $\pi_{0}$-FAST~\cite{pertsch2025fast} & 62.5\% & 29.1\% & 58.5\% & 21.9\% & 54.0\% & 10.8\% & 83.3\% & 66.6\% & 48.3\% \\
        \rowcolor[HTML]{EFEFEF} \ours & 73.8\% & \textbf{58.0\%} & \textbf{83.3\%} & \textbf{66.7\%} & \textbf{83.2\%} & \textbf{45.5\%} & \textbf{97.8\%} & \textbf{72.8\%} & \textbf{72.6\%} \\
        \midrule
        \multicolumn{11}{c}{\textbf{SimplerEnv on Google Robot Tasks}} \\
        \midrule
        \multirow{2}{*}{Model} & \multicolumn{4}{c}{\textbf{Visual Matching}} & & \multicolumn{4}{c}{\textbf{Variant Aggregation}} & \\
        \cline{2-5} \cline{7-10}
        & Pick Coke Can & Move Near & \begin{tabular}[c]{@{}l@{}}Open/Close Drawer\end{tabular} & \#Average & & Pick Coke Can & Move Near & \begin{tabular}[c]{@{}l@{}}Open/Close Drawer\end{tabular} & \#Average \\
        \cmidrule{1-10}
        RT-1~\citep{brohan2022rt} (Begin) & 2.7\% & 5.0\% & 13.9\% & 7.2\% & & 2.2\% & 4.0\% & 6.9\% & 4.4\% \\
        RT-1~\citep{brohan2022rt} ($15\%$) & 71.0\% & 35.4\% & 56.5\% & 54.3\% & & 81.3\% & 44.6\% & 26.7\% & 56.2\% \\
        RT-1~\citep{brohan2022rt} (Converged) & 85.7\% & 44.2\% & \textbf{73.0\%} & 74.6\% & & 89.8\% & 50.0\% & 32.3\% & 63.3\% \\
        RT-1-X~\cite{o2024open} & 56.7\% & 31.7\% & 59.7\% & 53.4\% & & 49.0\% & 32.3\% & 29.4\% & 39.6\% \\
        RT-2-X~\cite{o2024open} & 78.7\% & 77.9\% & 25.0\% & 60.7\% & & 82.3\% & 79.2\% & 35.3\% & 65.6\% \\
        Octo-Base~\cite{team2024octo} & 17.0\% & 4.2\% & 22.7\% & 16.8\% & & 0.6\% & 3.1\% & 1.1\% & 1.1\% \\
        OpenVLA~\cite{kim2024openvla} & 16.3\% & 46.2\% & 35.6\% & 27.7\% & & 54.5\% & 47.7\% & 17.7\% & 39.8\% \\
        TraceVLA~\cite{zheng2024tracevla} & 28.0\% & 53.7\% & 57.0\% & 42.0\% & & 60.0\% & 56.4\% & 31.0\% & 45.0\% \\
        RoboVLM~\cite{li2023generalist} & 77.3\% & 61.7\% & 43.5\% & 63.4\% & & 75.6\% & 60.0\% & 10.6\% & 51.3\% \\
        SpatialVLA~\cite{qu2025spatialvlaexploringspatialrepresentations} & 86.0\% & 77.9\% & 57.4\% & 73.8\% & & 88.0\% & 72.7\% & 41.8\% & 70.7\% \\
        HPT~\citep{wang2024scaling} & 56.0\% & 60.0\% & 24.0\% & 46.0\% & & -------------- & -------------- & -------------- & -------------- \\
        $\pi_{0}$~\cite{black2024pi_0} & 72.7\% & 65.3\% & 38.3\% & 58.8\% & & 75.2\% & 63.7\% & 25.6\% & 54.8\% \\
        $\pi_{0}$-FAST~\cite{pertsch2025fast} & 75.3\% & 67.5\% & 42.9\% & 61.9\% & & 77.6\% & 68.2\% & 31.3\% & 59.0\% \\
        \rowcolor[HTML]{EFEFEF} \ours & \textbf{97.0\%} & \textbf{80.4\%} & 58.8\% & \textbf{78.7\%} & & \textbf{98.0\%} & \textbf{79.7\%} & \textbf{44.6\%} & \textbf{74.1\%} \\
        \bottomrule
    \end{tabular}
    }
    \label{tab:simplerenv_combined}
     \vspace{-3ex}
\end{table*}

\subsection{\textbf{Real-World Embodiment Control}}
\label{sec:realworld}

\begin{figure*}[h]
    \vspace{-2ex}
    \begin{center}
        \includegraphics[width=\linewidth]{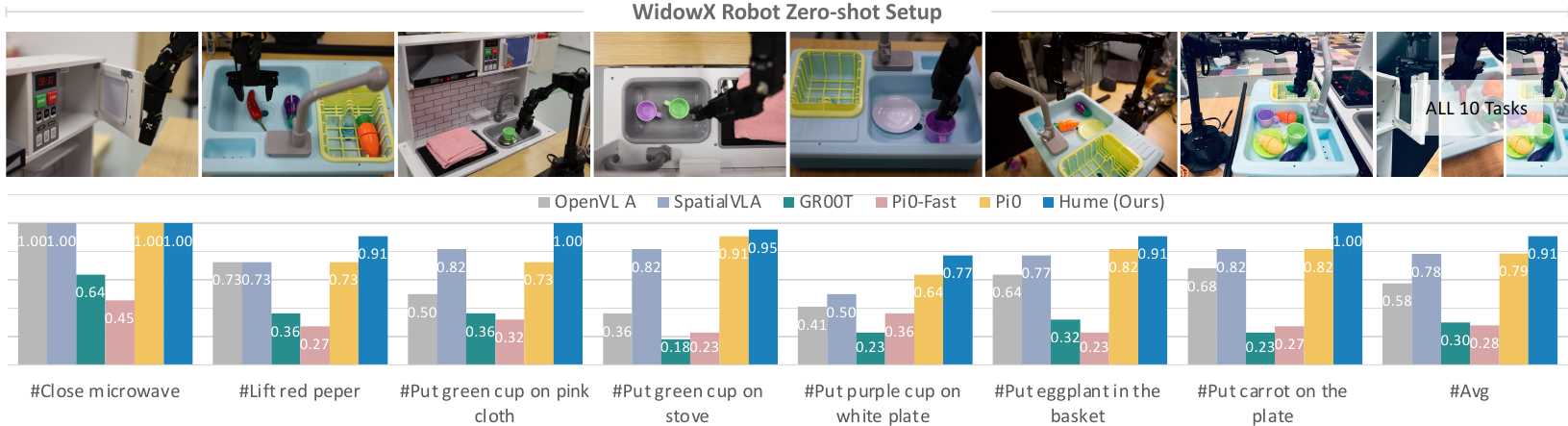}
    \end{center}
    \vspace{-1.5ex}
    \caption{\textbf{Real-world evaluation on WidowX Robot tasks.} We evaluate \ours across 10 tasks with varying backgrounds, poses, and motion distractors. \ours achieves the highest average success rate, surpassing $\pi_0$ and all other generalist manipulation policies in comparative evaluations.}
    \label{fig:widowx_real}
    \vspace{-2ex}
\end{figure*}

\textbf{Real-world WidowX Evaluation.} \cref{fig:widowx_real} presents the results of the real-world evaluation in WidowX robot platform.
We compared some representative single-system VLA models and dual-system VLA models on multiple tasks. We observe that, in simple task scenarios (\#1-2), most policies exhibit some generalizability, successfully completing tasks in unseen environments. 
However, in more complex tasks (\#3-7), policies such as GR00T, $\pi_0$-FAST, and OpenVLA struggle with manipulation, frequently encountering grasp failures issues, such as inability to accurately grasp or place target objects. Even when policies attempt to recover from failures, they often trap in error states that prevent successful execution.
In contrast, \ours leverages value-guided thinking to effectively recover from failures, demonstrating superior performance on various complex tasks. 
That is, \textit{when \ours falls into a wrong state, it can evaluate multiple candidate actions and select another correct trajectory forward.} As a result, even if failures occur during initial attempts, \ours can adjust its trajectory and successfully complete the task on subsequent attempts (please refer \href{https://hume-vla.github.io}{https://hume-vla.github.io} for more details), achieving strong robustness ability across various complex unseen tasks (91\% average success rate), improves by + 12\% over $\pi_0$ and + 33\% over OpenVLA.

\begin{figure*}[t]
    \begin{center}
        \includegraphics[width=\linewidth]{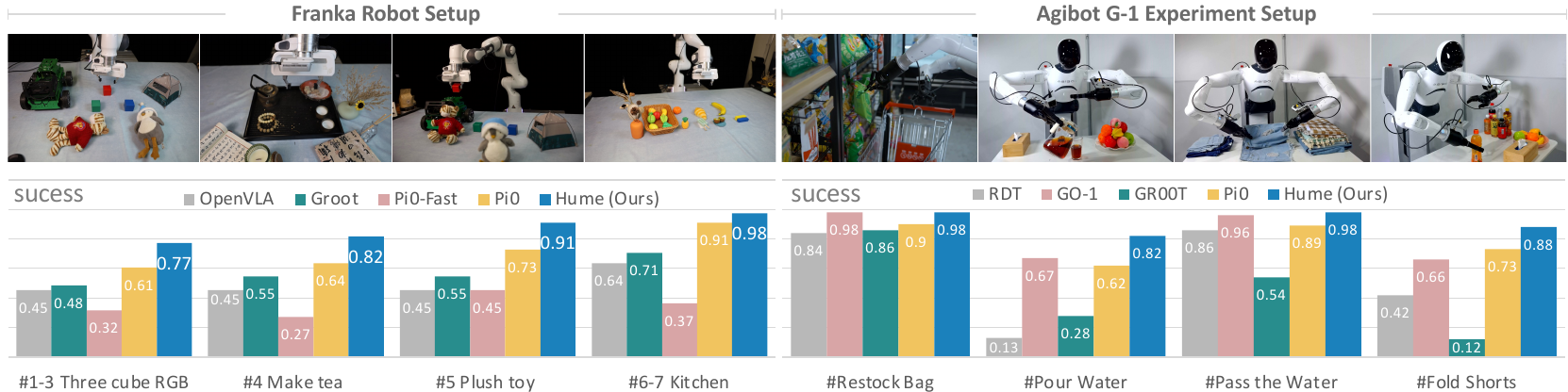}
    \end{center}
    \vspace{-1.5ex}
    \caption{\textbf{Evaluation on Franka and Agibot G-1 Robot.} We evaluate \ours across 11 real-world common tasks on Franka and Agibot G-1 robot.}
    \label{fig:franka_agibot_real}
    \vspace{-2.5ex}
\end{figure*}

\textbf{Real-world Franka and Agibot G-1 Evaluation.} \cref{fig:franka_agibot_real} presents the results of the real-world evaluation on Franka and Agibot G-1 robot platforms. The task design incorporates multiple real-world long-horizon tasks, deformable objects manipulation, tool-usage, and other challenging scenarios, with further validation conducted on both the Franka robot and the humanoid robot Agibot G-1. 
We observe that \textit{the long-term planning capability of System-2 thinking employed by \ours helps us better solve long-horizon tasks.} For example, in Agibot's long-horizon deformable objects manipulation task (\#Fold Shorts), where the models need to make the robot fold different shorts, \ours achieves a success rate of 88\%, improving by +15\% over $\pi_0$. In the complex long-horizon task (\#Pour Water), Hume achieves success rate of 82\%, significantly improving by +20\% over $\pi_0$, and +60\% over GR00T. Additionally, \ours also achieves an average success rate of 87\% across various tasks on the Franka robot, improving by +14.75\% over $\pi_0$, and +37.25\% over OpenVLA.

\subsection{\textbf{Ablations on Design Decisions}}
\begin{figure*}[b]
    \begin{center}
        \includegraphics[width=\linewidth]{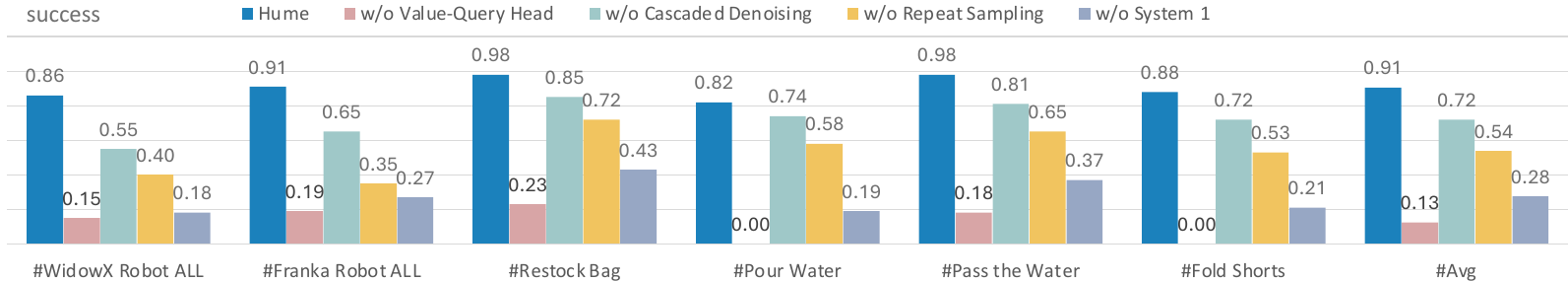}
    \end{center}
    \vspace{-1.5ex}
    \caption{\textbf{Real-world Ablations on WidowX, Franka and Agibot G-1 Robot.} We conducted ablation studies of \ours across 3 different real-world robotic platforms, covering 15 robot learning scenarios and 21 real-world manipulation tasks.}
    \label{fig:franka_agibot_real_ablation}
    \vspace{-2ex}
\end{figure*}

In this section, we conduct value-guided thinking and cascaded denoising ablations across multiple tasks in both simulation and real-world environments , with results presented in~\cref{tab:ablation_fintuning} and~\cref{fig:franka_agibot_real_ablation}.

\textbf{Effect of Cascaded Denoising.} 
According to the ablation results (\#1\textit{v.s.}\#2), the proposed cascaded denoising \textit{employs System 1 to remove the remaining noise, enabling the robot to perform precise and dexterous movements.}
Models w/o cascaded denoising utilize System 2 to complete the entire denoising process, leading to all candidates being sampled from the same distribution. This consequently reduces the range of candidates that the model choose from, resulting in suboptimal candidates selection. The models suffer a significant performance drop in variant aggregation, showing an average decline of -3.2\% across multiple SimplerEnv tasks, -2.7\% across LIBERO tasks, and -19\% in real-world robot tasks.

Furthermore, we compared \ours with the model directly using the optimal candidate generated by System 2 (\#1\textit{v.s.}\#4). Without System 1 to remove the remaining noise, the model w/o System 1 cannot perform precise and dexterous movements, ultimately leading to an average decrease of -9.8\% across SimplerEnv tasks, -8.8\% across LIBERO tasks, and -63\% in real-world robot tasks.

\textbf{Effect of Value-Guided Thinking.} 
According to the ablation results (\#1\textit{v.s.}\#5), the proposed value-guided thinking 
\textit{enables the System 2 to select the most valuable candidate from multiple noisy candidates to pass to System 1, which effectively improves \ours's performance in handling complex robot control tasks.}
Models w/o value-guided thinking randomly select 1 out of 5 candidate actions generated by System 2 to pass to System 1. Since there are candidates with different levels of noise in the candidate actions, this random selection strategy may select harmful candidates for System 1, ultimately resulting in a significant decrease in success rate, showing an average decline of -14.95\% across multiple SimplerEnv tasks, -13.7\% across LIBERO tasks. Notably, this performance degradation is even more pronounced in more complex real-world robot scenarios, with an astonishing average decrease of -78\% across various tasks in real-world environments.

Additionally, according to the ablation results (\#1\textit{v.s.}\#3), we also compared \ours with model generate only one candidate from System 2 then passing it to System 1. Models w/o repeat sampling, due to having no additional candidates to choose from, cannot leverage the estimated value to provide more helpful action selection. This also led to performance degradation, with an average decrease of -6.2\% across SimplerEnv tasks, -4.8\% across LIBERO tasks, and -37\% in real-world robot tasks.

\label{sec:ablation}
\begin{table*}[!t]
\centering
\caption{\textbf{Ablations in LIBERO and SimplerEnv tasks.} We conducted ablation studies of \ours across LIBERO~\citep{liu2023libero} and SimplerEnv~\citep{li24simpler} on WidowX and Google Robot tasks. Models are trained with mixtures from the OXE dataset~\citep{o2024open} in the SimplerEnv experiments.}
\resizebox{\textwidth}{!}{
\begin{tabular}{l|cccccccccc}
\toprule
\multicolumn{11}{c}{\textbf{LIBERO Tasks}} \\
\midrule
\#setting & \multicolumn{2}{c}{LIBERO-Spatial} & \multicolumn{2}{c}{LIBERO-Object} & \multicolumn{2}{c}{LIBERO-Goal} & \multicolumn{2}{c}{LIBERO-Long} & \multicolumn{2}{c}{Average} \\
\midrule
$[1]$. \ours & \multicolumn{2}{c}{98.6 $\pm$ 0.2\%} & \multicolumn{2}{c}{99.8 $\pm$ 0.1\%} & \multicolumn{2}{c}{99.4 $\pm$ 0.3\%} & \multicolumn{2}{c}{96.7 $\pm$ 0.9\%} & \multicolumn{2}{c}{98.6 $\pm$ 0.7\%} \\
$[2]$. w/o Cascaded Denoising & \multicolumn{2}{c}{95.4 $\pm$ 0.8\%} & \multicolumn{2}{c}{97.2 $\pm$ 0.5\%} & \multicolumn{2}{c}{96.8 $\pm$ 0.6\%} & \multicolumn{2}{c}{94.2 $\pm$ 0.7\%} & \multicolumn{2}{c}{95.9 $\pm$ 0.5\%} \\
$[3]$. w/o Repeat Sampling & \multicolumn{2}{c}{93.6 $\pm$ 0.4\%} & \multicolumn{2}{c}{94.8 $\pm$ 0.2\%} & \multicolumn{2}{c}{95.2 $\pm$ 0.3\%} & \multicolumn{2}{c}{91.4 $\pm$ 0.9\%} & \multicolumn{2}{c}{93.8 $\pm$ 0.5\%} \\
$[4]$. w/o System 1 & \multicolumn{2}{c}{90.2 $\pm$ 0.6\%} & \multicolumn{2}{c}{91.8 $\pm$ 0.9\%} & \multicolumn{2}{c}{92.4 $\pm$ 0.7\%} & \multicolumn{2}{c}{84.6 $\pm$ 0.2\%} & \multicolumn{2}{c}{89.8 $\pm$ 0.6\%} \\
$[5]$. w/o Value-Query Head & \multicolumn{2}{c}{85.2 $\pm$ 0.2\%} & \multicolumn{2}{c}{86.9 $\pm$ 0.4\%} & \multicolumn{2}{c}{88.2 $\pm$ 0.5\%} & \multicolumn{2}{c}{79.4 $\pm$ 0.6\%} & \multicolumn{2}{c}{84.9 $\pm$ 0.5\%} \\
\midrule
\multicolumn{11}{c}{\textbf{SimplerEnv on WidowX Robot Tasks}} \\
\midrule
\multirow{2}{*}{\#setting} & \multicolumn{2}{c}{Put Spoon on Towel} & \multicolumn{2}{c}{Put Carrot on Plate} & \multicolumn{2}{c}{Stack Green Block on Yellow Block} & \multicolumn{2}{c}{Put Eggplant in Yellow Basket} & \multicolumn{2}{c}{Overall} \\
\cmidrule{2-11}
 & Grasp Spoon & Success & Grasp Carrot & Success & Grasp Green Block & Success & Grasp Eggplant & Success & \multicolumn{2}{c}{Average} \\
\midrule
$[1]$. \ours & 73.8\% & 58.0\% & 83.3\% & 66.7\% & 83.2\% & 45.5\% & 97.8\% & 72.8\% & \multicolumn{2}{c}{72.6\%} \\
$[2]$. w/o Cascaded Denoising & 70.2\% & 55.6\% & 78.1\% & 62.5\% & 79.6\% & 42.1\% & 93.9\% & 67.3\% & \multicolumn{2}{c}{68.7\%} \\
$[3]$. w/o Repeat Sampling & 68.8\% & 49.2\% & 76.8\% & 57.9\% & 75.4\% & 39.9\% & 90.2\% & 66.9\% & \multicolumn{2}{c}{65.6\%} \\
$[4]$. w/o System 1 & 64.7\% & 42.3\% & 74.3\% & 53.4\% & 71.2\% & 36.2\% & 87.3\% & 63.1\% & \multicolumn{2}{c}{61.6\%} \\
$[5]$. w/o Value-Query Head & 58.2\% & 36.8\% & 67.6\% & 47.3\% & 66.9\% & 31.9\% & 83.6\% & 57.8\% & \multicolumn{2}{c}{56.3\%} \\
\midrule
\multicolumn{11}{c}{\textbf{SimplerEnv on Google Robot Tasks}.} \\
\midrule
\multirow{2}{*}{\#setting} & \multicolumn{5}{c}{Visual Matching} & \multicolumn{5}{c}{Variant Aggregation} \\
\cmidrule(lr){2-6} \cmidrule(lr){7-11}
 & Pick Coke Can & Move Near & Open/Close Drawer & \multicolumn{2}{c}{Average} & Pick Coke Can & Move Near & Open/Close Drawer & \multicolumn{2}{c}{Average} \\
\midrule
$[1]$. \ours & 97.0\% & 80.4\% & 58.8\% & \multicolumn{2}{c}{78.7\%} & 98.0\% & 79.7\% & 44.6\% & \multicolumn{2}{c}{74.1\%} \\
$[2]$. w/o Cascaded Denoising & 95.3\% & 75.4\% & 57.5\% & \multicolumn{2}{c}{76.1\%} & 94.8\% & 77.8\% & 42.2\% & \multicolumn{2}{c}{71.6\%} \\
$[3]$. w/o Repeat Sampling & 94.0\% & 70.8\% & 54.2\% & \multicolumn{2}{c}{73.0\%} & 92.2\% & 74.6\% & 39.3\% & \multicolumn{2}{c}{68.7\%} \\
$[4]$. w/o System 1 & 92.4\% & 65.4\% & 52.8\% & \multicolumn{2}{c}{70.2\%} & 89.9\% & 70.8\% & 35.9\% & \multicolumn{2}{c}{65.5\%} \\
$[5]$. w/o Value-Query Head & 89.9\% & 63.8\% & 48.9\% & \multicolumn{2}{c}{67.5\%} & 85.4\% & 65.9\% & 30.2\% & \multicolumn{2}{c}{60.5\%} \\
\bottomrule
\end{tabular}
}
\label{tab:ablation_fintuning}
\vspace{-2.5ex}
\end{table*}

\section{Conclusion and Limitations}
In this paper, we present \ours, a dual-system Vision-Language-Action (VLA) model to explore human-like thinking capabilities for generalist robot policy. \ours implements value-guided System-2 thinking by performing effective best-of-N selection with state-action estimation, and integrate System 1\&2 with the proposed cascaded action denoising to achieve rapid and fluid control for dexterous tasks.
With extensive experiments in both simulation and real robot platforms, we validated that \ours outperforms current state-of-the-art models, demonstrating the superiority of the \ours on various robot tasks, especially when failures occur in complex tasks during the deployment time, showing a promising research direction on generalist robot policy.

\textbf{Limitations.}
First, the value-guided System-2 thinking is limited by the quality of sampled candidate action chunks, 
how to include more high-value candidates among the sampled candidates is a question worthy of discussion.
Second, the estimated state-action value is not well-aligned with semantics, remaining a research direction for better value learning.
Last, the System-2 thinking paradigm implemented in \ours is still relatively naive, and future work can explore more sophisticated paradigms such as tree search, self-correction, and reinforcement learning approaches.
\label{sec:conclusion}

\newpage
\appendix
\section{\ours Visualization Analysis}
\label{sec:visualization}
In this section, we first demonstrate the detailed inference process of \ours through the dexterous Push-T task in \cref{sec:Hume_Inference_Visualization}, then visualize two key designs of \ours: value-guided System-2 thinking and cascaded action denoising in \cref{sec:Value-Guided_Thinking_Visualization} and \cref{sec:Cascaded_Action_Denoising_Visualization} to provide a comprehensive understanding of \ours.

\subsection{Hume Workflow Visualization}
\label{sec:Hume_Inference_Visualization}
To illustrate the detailed inference workflow of \ours, we evaluate \ours on the Push-T task that needs complex and contact-rich controls to push the T block precisely.
The Push-T task requires the policy to control a blue dot on a two-dimensional plane to push a gray T-shaped block into the green area. Since the action space is two-dimensional, we can visualize the actions predicted by \ours as trajectories on a plane.
\begin{figure*}[h]
    \vspace{-3ex}
    \begin{center}
        \includegraphics[trim=0 0 0 0, clip, width=0.98\textwidth]{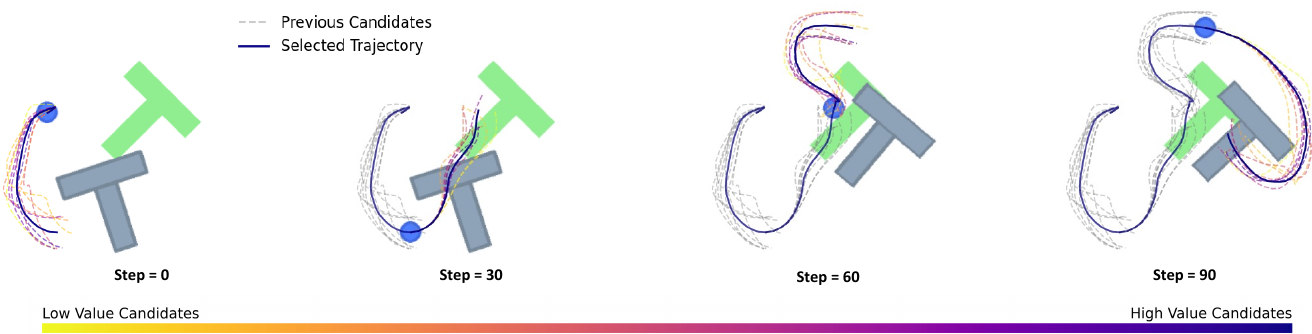}
    \end{center}
    \vspace{-1.5ex}
    \caption{\textbf{Visualization of Hume in Push-T.} We visualize the candidate actions $\mathbf{A}_{t}^{\tau_i}$ sampled from System 2 with dashed lines and the final executed action $\Tilde{\mathbf{A}}_{t+kh}^{1}$ from System 1 with solid line. The intensity of colors of lines indicates the magnitude of state-action values $Q(\mathbf{q_t}, \mathbf{A}_{t}^{\tau_i})$ of candidates.}
    \label{fig:pusht}
    \vspace{-3ex}
\end{figure*}

As shown in \cref{fig:pusht}, we illustrate the detailed inference process of \ours in the Push-T task. Specifically, in inference, we sample the candidate actions $\mathbf{A}_{t}^{\tau_i}$ from System 2 at time steps $t=0, 30, 60, 90$ with a horizon of $H=30$, and produce 10 candidate actions at each timestep. The selected action with highest value is conveyed to System 1 and continuously denoised to executed action as drew with solid line. We can see that the final denoised action from System 1 is smoother and more delicate for accomplishing the task.

\subsection{Value-Guided Thinking Visualization}
\label{sec:Value-Guided_Thinking_Visualization}
\cref{fig:pca_map} visualizes estimated state-action values of the candidate actions at different time steps in the LIBERO-GOAL benchmark.
Since the action space in LIBERO is 7-dimensional, we used Principal Component Analysis (PCA) to project the high dimensional actions onto a two-dimensional plane.
For each projected candidate action, we use different colors to represent their corresponding state-action values estimated by the value-query head.
These projected points together with their corresponding state-action values generate the value map shown in the figure.
In the value map, yellow color represent actions with high state-action values, while purple color represent actions with low state-action values.
Additionally, we also show the ground truth actions from collected demonstrations in the value map for comparison.


By observing the positions of ground truth actions in the value map, we find these ground truth actions are consistently located in high-value regions, which demonstrates that System 2's value-query head is capable of making reasonable estimates of the state-action values.
We also find ground truth actions are not located at the highest-value positions in the value map, which proves System 2's value-query head has not been overfitted to ground truth actions, but is able to estimate appropriate state-action values in whole action spaces, guiding System 2 to select the optimal candidate action.
Furthermore, by comparing value maps across different timesteps, we observe adjacent timesteps (step=0 and step=1) have similar value maps, while value maps at distant timesteps (step=50 and step=100) exhibit significant differences. 
This demonstrates the value-query head can reasonably adjust its estimation of state-action values by capturing nature world dynamics, guiding System 2 to make smooth choices for robot control.

\subsection{Cascaded Action Denoising Visualization}
\label{sec:Cascaded_Action_Denoising_Visualization}
\begin{figure*}[h]
    \begin{center}
        \includegraphics[trim=0 0 0 0, clip, width=1.0\textwidth]{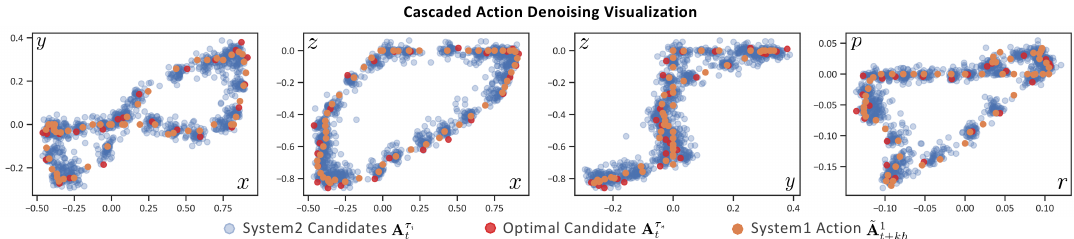}
    \end{center}
    \vspace{-1.5ex}
    \caption{\textbf{Cascaded Action Denoising on Different Action Dimension.} We visualize denoised actions grouped to coordinate pairs (X-Y, X-Z, Y-Z, and R-P) from 7-dimensional action space (X, Y, Z, Roll, Pitch, Yaw, Gripper). For each group, we display the candidate actions $\mathbf{A}_{t}^{\tau_i}$ sampled from System 2, the optimal candidate $\mathbf{A}_{t}^{\tau^*}$, and final action $\Tilde{\mathbf{A}}_{t+kh}^{1}$ denoised by System 1.}
    \label{fig:cascaded_denoise}
    \vspace{-2ex}
\end{figure*}

\cref{fig:cascaded_denoise} visualizes the cascaded action denoising process in LIBERO-OBJECT.
For the 7 dimensions in action space (X, Y, Z, Roll, Pitch, Yaw, Gripper), we pair them into combinations for illustration, \emph{i.e.}, X-Y, X-Z, Y-Z, and R-P.
The drew points is down-sampled from the actual denoised action sequences for accomplishing one task, where the blue points represent candidate actions $\mathbf{A}_{t}^{\tau_i}$ sampled from System 2's denoising head, the red points represent optimal candidates $\mathbf{A}_{t}^{\tau^*}$ selected by System 2, and the orange points represent final actions $\Tilde{\mathbf{A}}_{t+kh}^{1}$ denoised by System 1.
The red points and orange points are generally distributed within the region covered by blue points, while the distribution of orange points shows slight differences from the red points. This demonstrates that cascaded action denoising is actually refining selected actions from System 2 with higher-frequency new observation inputs in System 1, achieving accurate, fluid, and delicate robot control.

\section{Implementation Details}
\label{sec:impl_detail}
In this section, we provide further implementation details of \ours, including the training details of the value-query head and the hyperparameters used by the model during training.
\subsection{Value Objective Functions}
Formally, the goal of the value-query Head is to learn $Q_\theta(\mathbf{q}_t, \mathbf{A}_t)$, which is the optimal estimate~\citep{liu2023improving} of the state-action value function ${Q^\pi(\mathbf{q}_t, \mathbf{A}_t) = {\frac{1}{1-\gamma}\sum_{t} \mathbb{E}_{\mathbf{A}_t \sim \pi(\mathbf{q}_t)}\left[\gamma^t r(\mathbf{q}_t, \mathbf{A}_t)|\mathbf{q}_{t_0}=\mathbf{q}_{t},\mathbf{A}_{t_0}=\mathbf{A}_t\right]}}$ in a Markov Decision Process $\mathcal{M} = (\mathcal{S}, \mathcal{A}, P, r, \rho, \gamma)$. Here $\mathcal{S}, \mathcal{A}$ denote the state and action spaces, while $P(\mathbf{q}' | \mathbf{q}, \mathbf{A})$ and $r(\mathbf{q}_t, \mathbf{A}_t)$ are the dynamics and reward functions. $\rho(\mathbf{q})$ denotes the initial state distribution, and $\gamma \in (0,1)$ denotes the discount factor. 
The training objective of the Value-Query Head is to minimize the Bellman error with a regularization term $\mathcal{R}(\theta)$, which is defined as:
\begin{align}
    \label{eq:cql_training_supp}
    \!\!\!\!\!\min_\theta {\alpha\mathcal{R}(\theta)} + \frac{1}{2} {\mathbb{E}_{\mathbf{q}_t, \mathbf{A}_t, \mathbf{q}_t^\prime\sim \mathcal{D}}\left[\left(Q_\theta(\mathbf{q}_t, \mathbf{A}_t) - \mathcal{B}^\pi\bar{Q}(\mathbf{q}_t, \mathbf{A}_t)\right)^2 \right]},
\end{align}

The second term in~\cref{eq:cql_training_supp} is the standard TD error~\cite{lillicrap2015continuous,fujimoto2018addressing,haarnoja2018sacapps}, where $Q_\theta(\mathbf{q}_t, \mathbf{A}_t)$ is the output of the value-query head, and $\mathcal{B}^\pi \bar{Q} (\mathbf{q}_t, \mathbf{A}_t)$ is the backup operator applied to the delayed target Q-network $\bar{Q}$: $\mathcal{B}^\pi \bar{Q}(\mathbf{q}_t, \mathbf{A}_t) := r(\mathbf{q}_t, \mathbf{A}_t) + \gamma \mathbb{E}_{\mathbf{A}_t^\prime \sim \pi(\mathbf{A}_t^\prime|\mathbf{q}_t^\prime)}[\bar{Q}(\mathbf{q}_t^\prime, \mathbf{A}_t^\prime)]$, which can be calculated from the offline dataset $\mathcal{D} = \{(\mathbf{q}_t, \mathbf{A}_t, r, \mathbf{q}_t^\prime)\}$. 
The $\mathcal{R}(\theta)$ in~\cref{eq:cql_training_supp} is a calibrated conservative regularizer that aims to prevent overestimation in the Q-values for OOD actions by penalizing the Q-values, and compensating for this pessimism on actions seen in the training dataset, and $\alpha$ is a hyperparameter to control the conservative penalty. Specifically, the regularization term $\mathcal{R}(\theta)$ is defined as:

\begin{align}
\mathcal{R}(\theta) := \mathbb{E}_{\mathbf{q}_t \sim \mathcal{D}, a \sim \pi}\left[\max \left(Q_{\theta}(\mathbf{q}_t, \mathbf{A}_t), Q^{\mu}(\mathbf{q}_t, \mathbf{A}_t)\right)\right]-\mathbb{E}_{\mathbf{q}_t, \mathbf{A}_t \sim \mathcal{D}}\left[Q_{\theta}(\mathbf{q}_t, \mathbf{A}_t)\right]
\end{align}

where $Q^{\mu}(\mathbf{q}_t, \mathbf{A}_t)$ is the value function of the calibrated policy $\mu$.
\subsection{Training and Inference Hyperparameters}
\textbf{LIBERO}
In LIBERO, \ours takes images of third-person camera, wrist camera, and robot state as input. We set the chunk size of System 2's output to 16 and the chunk size of System 1's output to 8. The model was trained using 8 GPUs with a batch size of 16.

\textbf{SimplerEnv}
In SimplerEnv, \ours takes image of third-person camera and robot state as input. In Bridge tasks, we set the chunk size of System 2's output to 8 and the chunk size of System 1's output to 4. In Google Robot tasks, we set the chunk size of System 2's output to 4 and the chunk size of System 1's output to 2. The model was trained using 8 GPUs with a batch size of 32.

\textbf{Franka}
In real-world experiment on Franka-Emika-Panda, \ours takes images of third-person camera and robot state as input. We set the chunk size of System 2's output to 16 and the chunk size of System 1's output to 8. The model was trained using 4 GPUs with a batch size of 32.

\textbf{Widowx}
In real-world experiment on Widowx 250s, \ours uses the same training settings as in the simulation environment of Bridge in SimplerEnv, which takes images of third-person camera and robot state as input, and the chunk size of System 2 is 8 and the chunk size of System 1 is 4, trained using 8 GPUs with a batch size of 32.

\textbf{Agibot G-1}
In real-world experiment on Agibot G-1, \ours takes images from the head camera and wrist cameras on both arms along with robot state as input. We set the chunk size of System 2's output to 30 and the chunk size of System 1's output to 15. The model was trained using 8 GPUs with a batch size of 8.

\section{Experiment Details}
\label{sec:exp_detail}
In this section, we will provide experiment details including evaluation setup and additional test results. 
In~\cref{sec:sim_benchmark}, we provide detailed descriptions of test setup for standard simulation benchmark and the implementation details of all comparison methods along with the detailed test results.
In~\cref{sec:real_world_eval}, we introduce the detailed setup of test tasks on real robot platforms and testing standards, and provide the detailed test results.
\subsection{Simulation Benchmark Details}
\label{sec:sim_benchmark}
\textbf{Simulation Benchmark Setup.}
In LIBERO, all methods use third-person camera, wrist camera, and robot state as input, where the results of Diffusion Policy and OpenVLA-OFT are from the technical report of OpenVLA-OFT~\cite{kim2025fine}, the results of $\pi_{0}$~\cite{black2024pi_0} and $\pi_0$-FAST~\cite{pertsch2025fast} are provided by Physical Intelligence's open-source repository, and the results of GR00T N1 are obtained by training and testing on the LIBERO dataset using its open source code.
In SimplerEnv, test results of RT-1~\citep{brohan2022rt}, RT-1-X~\cite{o2024open}, RT-2-X~\cite{o2024open}, Octo~\cite{team2024octo}, OpenVLA~\cite{kim2024openvla}, HPT~\citep{wang2024scaling}, TraceVLA~\cite{zheng2024tracevla}, RoboVLM~\cite{li2023generalist}, SpatialVLA~\cite{qu2025spatialvlaexploringspatialrepresentations} come from their official technical reports. The results of GR00T~\cite{bjorck2025gr00t}, $\pi_0$-FAST~\cite{pertsch2025fast}, and $\pi_{0}$~\cite{black2024pi_0} are obtained by us fine-tuning and testing them on the corresponding datasets using their official open-source code.

\textbf{Detailed Results of Simulation Benchmark.} 
Since the Google Robot tasks in SimplerEnv include various test settings such as environment layout, object position, and texture variations, we provide more detailed test results in~\cref{tab:simplerenv_google_robot_supp}.
\begin{table*}
    [t]
    \centering
    \caption{SimplerEnv evaluation results across different policies on Google
    Robot tasks.}
    \label{tab:simplerenv_google_robot_supp} \resizebox{\columnwidth}{!}{
    \begin{tabular}{l|lccccccccc}
        \toprule \multirow{5}{*}{\begin{tabular}[l]{@{}l@{}}Google Robot\\ Evaluation Setup\end{tabular}} & \multirow{5}{*}{Policy} & \multicolumn{4}{c}{\multirow{2}{*}{Pick Coke Can}}          & \multirow{2}{*}{Move Near}                                & \multicolumn{3}{c}{\multirow{2}{*}{Open / Close Drawer}} & \multirow{2}{*}{\begin{tabular}[c]{@{}c@{}}\#Overall\end{tabular}} \\
        \\
        \cmidrule{3-11}                                                                                   &                         & \begin{tabular}[c]{@{}c@{}}Horizontal\\ Laying\end{tabular} & \begin{tabular}[c]{@{}c@{}}Vertical\\ Laying\end{tabular} & \begin{tabular}[c]{@{}c@{}}Standing\end{tabular}         & \begin{tabular}[c]{@{}c@{}}Average\end{tabular}                   & \begin{tabular}[c]{@{}c@{}}Average\end{tabular} & \begin{tabular}[c]{@{}c@{}}Open\end{tabular} & \begin{tabular}[c]{@{}c@{}}Close\end{tabular} & \begin{tabular}[c]{@{}c@{}}Average\end{tabular} & \begin{tabular}[c]{@{}c@{}}Average\end{tabular} \\
        \midrule \multirow{14}{*}{\begin{tabular}[l]{@{}l@{}}Variant\\Aggregation\end{tabular}}           & RT-1 (begin)            & 2.2\%                                                       & 1.3\%                                                     & 03.1\%                                                   & 02.2\%                                                            & 04.0\%                                          & 00.5\%                                       & 13.2\%                                        & 06.9\%                                          & 04.4\%                                          \\
                                                                                                          & RT-1 ($15\%$)           & 92.0\%                                                      & 70.4\%                                                    & 81.3\%                                                   & 81.2\%                                                            & 44.6\%                                          & 21.2\%                                       & 32.3\%                                        & 26.8\%                                          & 50.9\%                                          \\
                                                                                                          & RT-1 (converged)        & 96.9\%                                                      & 76.0\%                                                    & 96.4\%                                                   & 89.8\%                                                            & 50.0\%                                          & 27.0\%                                       & 37.6\%                                        & 32.3\%                                          & 57.4\%                                          \\
                                                                                                          & RT-1-X                  & 56.9\%                                                      & 20.4\%                                                    & 69.8\%                                                   & 49.0\%                                                            & 32.3\%                                          & 06.9\%                                       & 51.9\%                                        & 29.4\%                                          & 36.9\%                                          \\
                                                                                                          & RT-2-X                  & 82.2\%                                                      & 75.4\%                                                    & 89.3\%                                                   & 82.3\%                                                            & 79.2\%                                          & 33.3\%                                       & 37.2\%                                        & 35.3\%                                          & 65.6\%                                          \\
                                                                                                          & Octo-Base               & 0.5\%                                                       & 00.0\%                                                    & 01.3\%                                                   & 00.6\%                                                            & 03.1\%                                          & 00.0\%                                       & 02.1\%                                        & 01.1\%                                          & 01.6\%                                          \\
                                                                                                          & OpenVLA                 & 71.1\%                                                      & 27.1\%                                                    & 65.3\%                                                   & 54.5\%                                                            & 47.7\%                                          & 15.8\%                                       & 19.5\%                                        & 17.7\%                                          & 40.0\%                                          \\
                                                                                                          & TraceVLA                & ---                                                         & ---                                                       & ---                                                      & 60.0\%                                                            & 56.4\%                                          & ---                                          & ---                                           & 31.0\%                                          & 49.1\%                                          \\
                                                                                                          & RoboVLM                 & 93.8\%                                                      & 49.8\%                                                    & 83.1\%                                                   & 75.6\%                                                            & 60.0\%                                          & 02.6\%                                       & 18.5\%                                        & 10.6\%                                          & 48.7\%                                          \\
                                                                                                          & SpatialVLA              & 93.3\%                                                      & 78.2\%                                                    & 92.4\%                                                   & 88.0\%                                                            & 72.7\%                                          & 28.6\%                                       & 55.0\%                                        & 41.8\%                                          & 67.5\%                                          \\
                                                                                                          & $\pi_{0}$               & 82.0\%                                                      & 58.0\%                                                    & 85.6\%                                                   & 75.2\%                                                            & 63.7\%                                          & 18.0\%                                       & 33.2\%                                        & 25.6\%                                          & 54.8\%                                          \\
                                                                                                          & $\pi_{0}$-FAST          & 84.0\%                                                      & 63.0\%                                                    & 85.8\%                                                   & 77.6\%                                                            & 68.2\%                                          & 24.0\%                                       & 38.6\%                                        & 31.3\%                                          & 59.0\%                                          \\
        \rowcolor[HTML]{EFEFEF}                                                                           & \ours                   & 99.0\%                                                      & 96.0\%                                                    & 99.0\%                                                   & 98.0\%                                                            & 79.7\%                                          & 38.0\%                                       & 51.2\%                                        & 44.6\%                                          & 74.1\%                                          \\
        \midrule \multirow{15}{*}{\begin{tabular}[l]{@{}l@{}}Visual\\Matching\end{tabular}}               & RT-1 (Begin)            & 5.0\%                                                       & 00.0\%                                                    & 03.0\%                                                   & 02.7\%                                                            & 05.0\%                                          & 00.0\%                                       & 27.8\%                                        & 13.9\%                                          & 07.2\%                                          \\
                                                                                                          & RT-1 ($15\%$)           & 86.0\%                                                      & 79.0\%                                                    & 48.0\%                                                   & 71.0\%                                                            & 35.4\%                                          & 46.3\%                                       & 66.7\%                                        & 56.5\%                                          & 54.3\%                                          \\
                                                                                                          & RT-1 (Converged)        & 96.0\%                                                      & 90.0\%                                                    & 71.0\%                                                   & 85.7\%                                                            & 44.2\%                                          & 60.1\%                                       & 86.1\%                                        & 73.1\%                                          & 67.7\%                                          \\
                                                                                                          & RT-1-X                  & 82.0\%                                                      & 33.0\%                                                    & 55.0\%                                                   & 56.7\%                                                            & 31.7\%                                          & 29.6\%                                       & 89.1\%                                        & 59.4\%                                          & 49.3\%                                          \\
                                                                                                          & RT-2-X                  & 74.0\%                                                      & 74.0\%                                                    & 88.0\%                                                   & 78.7\%                                                            & 77.9\%                                          & 15.7\%                                       & 34.3\%                                        & 25.0\%                                          & 60.5\%                                          \\
                                                                                                          & Octo-Base               & 21.0\%                                                      & 21.0\%                                                    & 09.0\%                                                   & 17.0\%                                                            & 04.2\%                                          & 00.9\%                                       & 44.4\%                                        & 22.7\%                                          & 14.6\%                                          \\
                                                                                                          & OpenVLA                 & 27.0\%                                                      & 03.0\%                                                    & 19.0\%                                                   & 16.3\%                                                            & 46.2\%                                          & 19.4\%                                       & 51.8\%                                        & 35.6\%                                          & 32.7\%                                          \\
                                                                                                          & TraceVLA                & ---                                                         & ---                                                       & ---                                                      & 28.0\%                                                            & 53.7\%                                          & ---                                          & ---                                           & 57.0\%                                          & 46.2\%                                          \\
                                                                                                          & RoboVLM                 & 94.0\%                                                      & 47.0\%                                                    & 91.0\%                                                   & 77.3\%                                                            & 61.7\%                                          & 33.3\%                                       & 53.1\%                                        & 43.2\%                                          & 60.7\%                                          \\
                                                                                                          & SpatialVLA              & 85.0\%                                                      & 76.0\%                                                    & 97.0\%                                                   & 86.0\%                                                            & 77.9\%                                          & 50.0\%                                       & 64.8\%                                        & 57.4\%                                          & 73.8\%                                          \\
                                                                                                          & HPT                     & ---                                                         & ---                                                       & ---                                                      & 56.0\%                                                            & 60.0\%                                          & ---                                          & ---                                           & 24.0\%                                          & 46.7\%                                          \\
                                                                                                          & $\pi_{0}$               & 76.0\%                                                      & 57.0\%                                                    & 85.1\%                                                   & 72.7\%                                                            & 65.3\%                                          & 30.0\%                                       & 46.6\%                                        & 38.3\%                                          & 58.8\%                                          \\
                                                                                                          & $\pi_{0}$-FAST          & 79.0\%                                                      & 61.0\%                                                    & 85.9\%                                                   & 75.3\%                                                            & 67.5\%                                          & 34.0\%                                       & 51.8\%                                        & 42.9\%                                          & 61.9\%                                          \\
        \rowcolor[HTML]{EFEFEF}                                                                           & \ours                   & 99.0\%                                                      & 93.0\%                                                    & 99.0\%                                                   & 97.0\%                                                            & 80.4\%                                          & 52.0\%                                       & 65.6\%                                        & 58.8\%                                          & 78.7\%                                          \\
        \bottomrule
    \end{tabular}
    }
\end{table*}

\subsection{Real-World Evaluation Details}
\label{sec:real_world_eval}
\textbf{Real-World Evaluation Setup.} 
In this section, we provide detailed descriptions of task setups on three real-world robot platforms: WidowX 250s, Franka-Panda-Emika, and Agibot G-1.
\begin{figure*}[t]
    \vspace{-2ex}
    \begin{center}
        \includegraphics[width=1.0\textwidth]{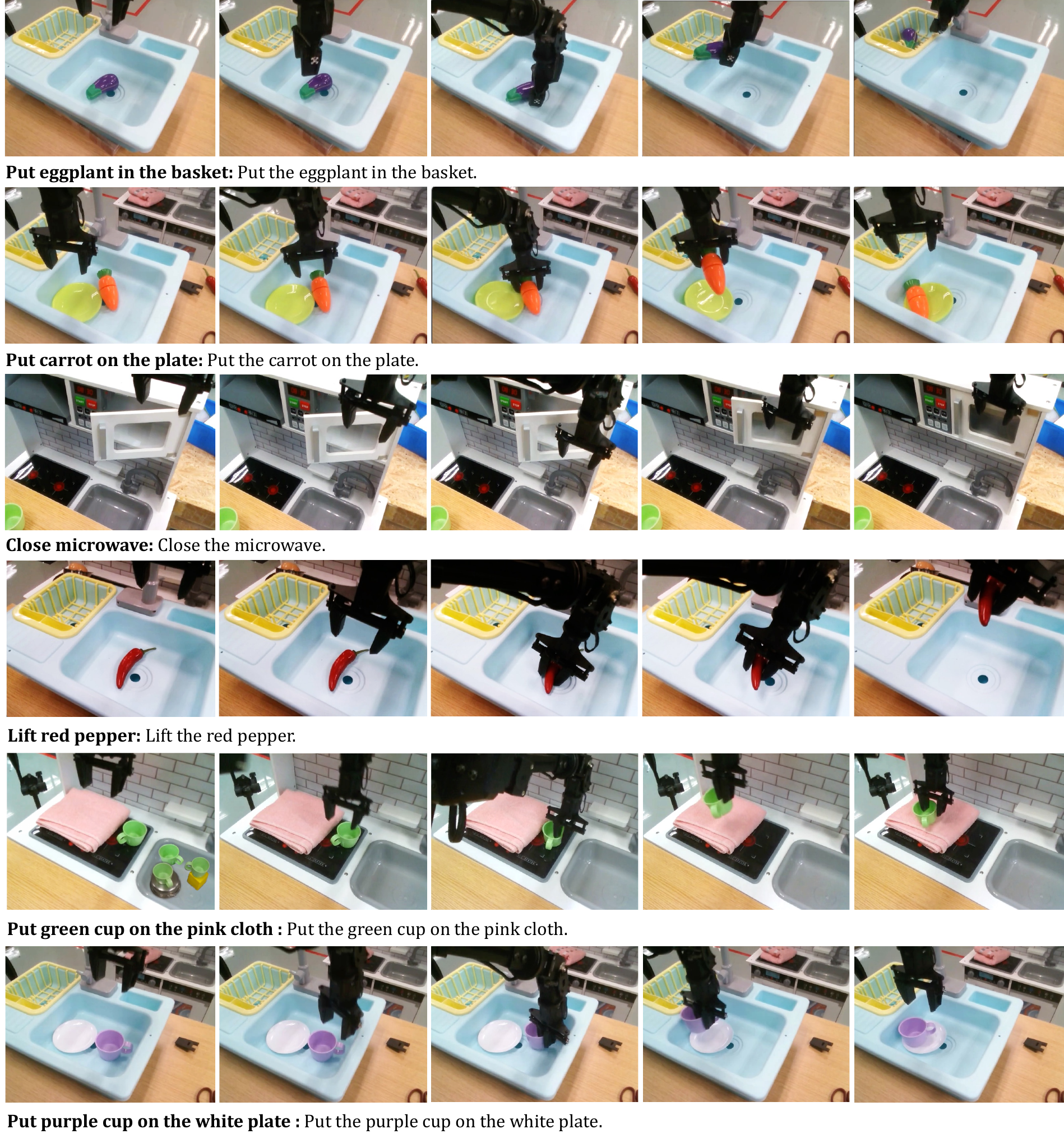}
    \end{center}
    \vspace{-1.5ex}
    \caption{\textbf{Evaluation Setup of WidowX 250s.} We evaluated models with 9 tasks on WidowX 250s to verify the model's learning ability on a large multi-task manipulation dataset.}
    \label{fig:supp_widowx_tasks}
    \vspace{-2ex}
\end{figure*}

As shown in~\cref{fig:supp_widowx_tasks}, the detailed task specifications on WidowX 250s are:
\begin{itemize}[leftmargin=8pt]
\item \textbf{Put eggplant in the basket}: A complex task requiring the robot
to identify and pick an eggplant from a sink containing multiple vegetables,
then place it in a yellow basket. This task evaluates object discrimination
and spatial awareness.

\item \textbf{Put carrot on the plate}: The robot needs to perform a pick-and-place
task by grasping a carrot from the sink and placing it on a plate,
assessing both grasping precision and placement accuracy.

    \item \textbf{Close microwave}: The robot must close a toy microwave door positioned
    at various angles (30°, 45°, 60°, and 90°), testing the model's capability
    to manipulate articulated objects in different configurations.

    \item \textbf{Lift red pepper}: A basic pick task requiring the robot to
    grasp and lift a red pepper from the sink, designed to evaluate the model's
    object localization accuracy.

    \item \textbf{Put green cup on the pink cloth}: This task suite comprises two
    scenarios testing vertical spatial understanding. In the first scenario, the
    robot grasps a green cup positioned either on a stove or elevated on a yellow
    block. In the second scenario, the cup is placed either at the bottom of a
    sink or elevated on a bowl. This variation in object heights challenges
    the model's ability to adapt its manipulation strategy according to
    spatial configurations.

    \item \textbf{Put purple cup on the white plate}: The robot must identify and transfer
    a purple cup to a white plate within a sink containing multiple plates,
    testing color recognition and precise manipulation.
 
 \end{itemize}

 \begin{figure*}[t]
    \vspace{-4ex}
    \begin{center}
        \includegraphics[width=1.0\textwidth]{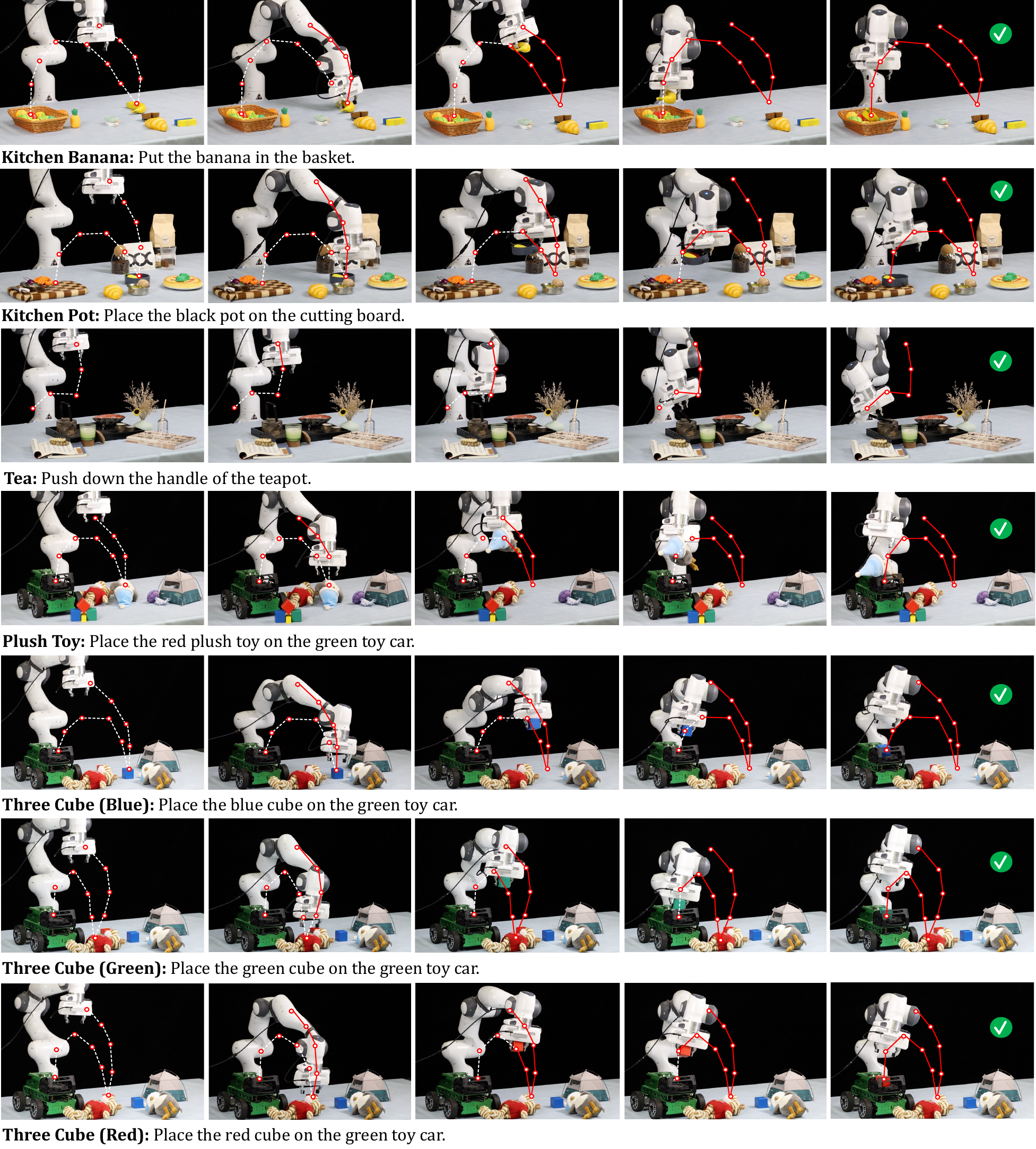}
    \end{center}
    \vspace{-1.5ex}
    \caption{\textbf{Evaluation Setup of Frank-Emika-Panda.} We evaluated policies on Fnraka robot with 7 tasks, including instruction following, articulated manipulation, and pick and place tasks.}
    \label{fig:supp_franka_tasks}
    \vspace{-2ex}
\end{figure*}

As shown in~\cref{fig:supp_franka_tasks}, the detailed task specifications on Frank-Emila-Panda are:
\begin{itemize}[leftmargin=8pt]
    \item \textbf{Kitchen Banana}: A pick-and-place task where
    the robot must transfer a banana from the table to a basket. With only 50
    human demonstrations, this task evaluates model performance with limited data.

    \item \textbf{Kitchen Pot}: Another pick-and-place task
    requiring the robot to grasp a white bowl from the right side of the table
    and place it on a cutting board, trained with 100 human demonstrations.

    \item \textbf{Tea}: The robot must push a teapot handle
    from a perpendicular to a a parallel position relative to the desktop,
    using its gripper tip. This task tests the manipulation of revolute joints
    and includes 50 human demonstrations.

   \item \textbf{Plush Toy}: The robot must identify and grasp the nearest plush toy among two options and place it on a green car. To rigorously assess spatial understanding, we systematically vary the relative positions of the plush toys during testing.

   \item \textbf{Three Cube}: An instruction-following
   task where the robot must identify and place a specifically colored cube (red,
   green, or blue) onto a green car. Training included 50 demonstrations for
   each color, totaling 150.

\end{itemize}

\begin{figure*}[t]
    \vspace{-4ex}
    \begin{center}
        \includegraphics[width=1.0\textwidth]{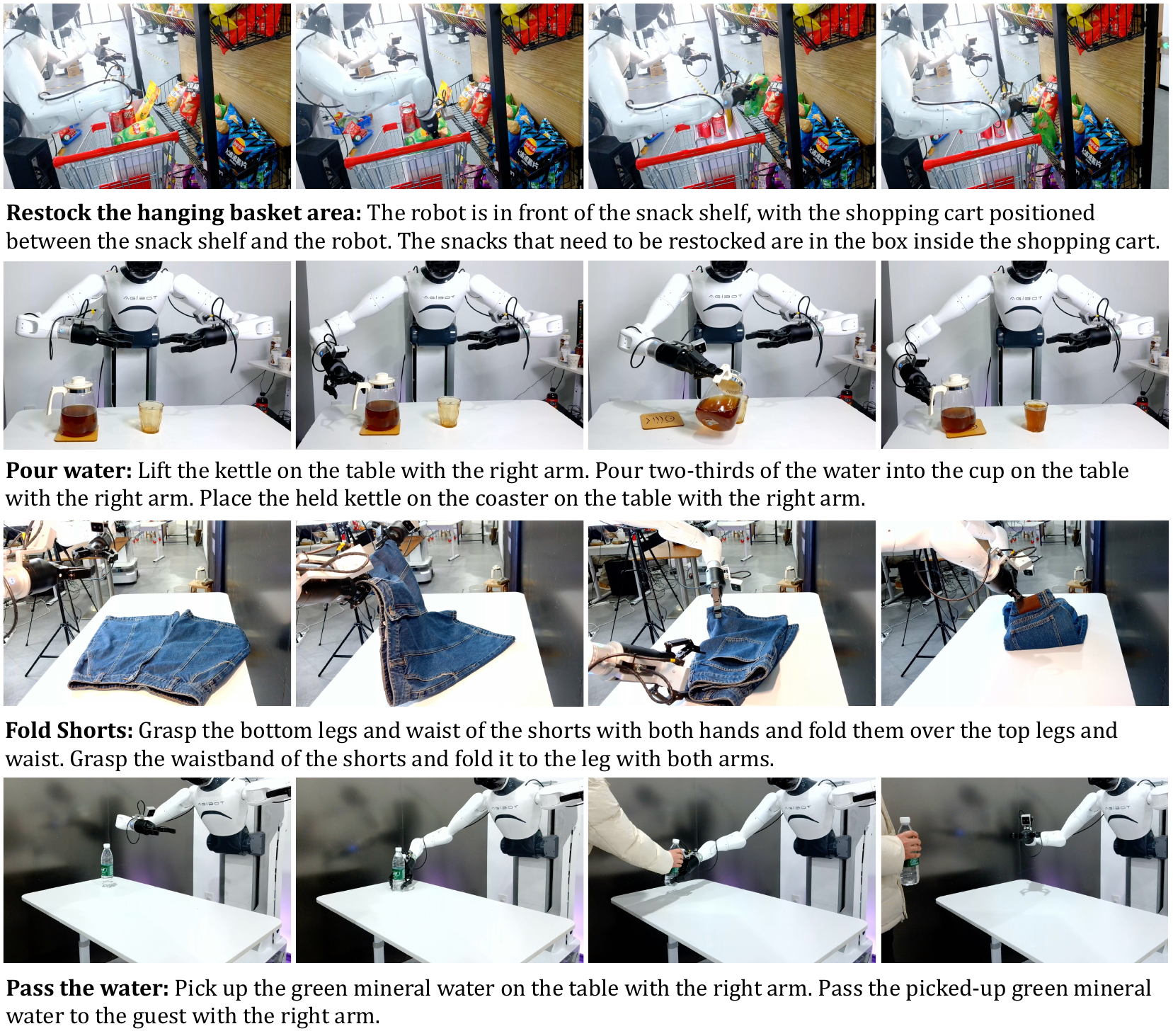}
    \end{center}
    \vspace{-1.5ex}
    \caption{\textbf{Evaluation Setup of AgiBot G-1.} We evaluated policies on four challenging tasks on AgiBot G-1 to test ability of controlling a humanoid robot completing dexterous and long-horizon tasks.}
    \label{fig:supp_agibot_tasks}
    \vspace{-2ex}
\end{figure*}

As shown in~\cref{fig:supp_agibot_tasks}, the detailed task specifications on AgiBot G-1 are:
\begin{itemize}[leftmargin=8pt]
    \item \textbf{Restock the hanging basket area}: This task requires the robot to grab snacks from a cart and place them at a designated location on the shelf. This task includes different types of snacks, different placements on the shelf, and interfering objects in the cart to verify the generalization of the model.

    \item \textbf{Pour water}: This is a long-horizon task that requires the robot to first grasp the handle of the teapot, lift the teapot and accurately pour the water from the teapot into the cup, and then put the teapot back on the mat after the cup is full. This task involves changing the material of the cup, as well as the position of the teapot and the cup. To complete this challenging task, the robot needs to accurately identify the location of the cup and pour water into it, and it needs to correctly sense the water level in the cup to avoid spilling water on the table.
 
    \item \textbf{Fold Shorts}: This is a long-horizon task that requires the collaboration of both arms and involves the manipulation of flexible objects. In this task, the robot first needs to accurately identify the position of the shorts and use the grippers on both arms to fold the shorts for the first time. After completing the first fold, the robot needs to confirm the current state of the shorts again and perform a second fold.
    During the testing of this task, we used shorts of different colors and materials to verify the model's long-horizon manipulation capabilities for flexible objects of different shapes.

    \item \textbf{Pass the water}: This task is designed to test the model's ability to follow instructions and collaborate with humans. In this task, the robot needs to grab the correct bottle according to the language instructions and hand it to the human.
    We used different types of bottles in the test, and we also arbitrarily adjusted the position of the bottles on the white table shown in the figure to verify the generalization of the robot to the position of objects.
 \end{itemize}
\begin{figure*}[t]
    \vspace{-2ex}
    \begin{center}
        \includegraphics[width=1.0\textwidth]{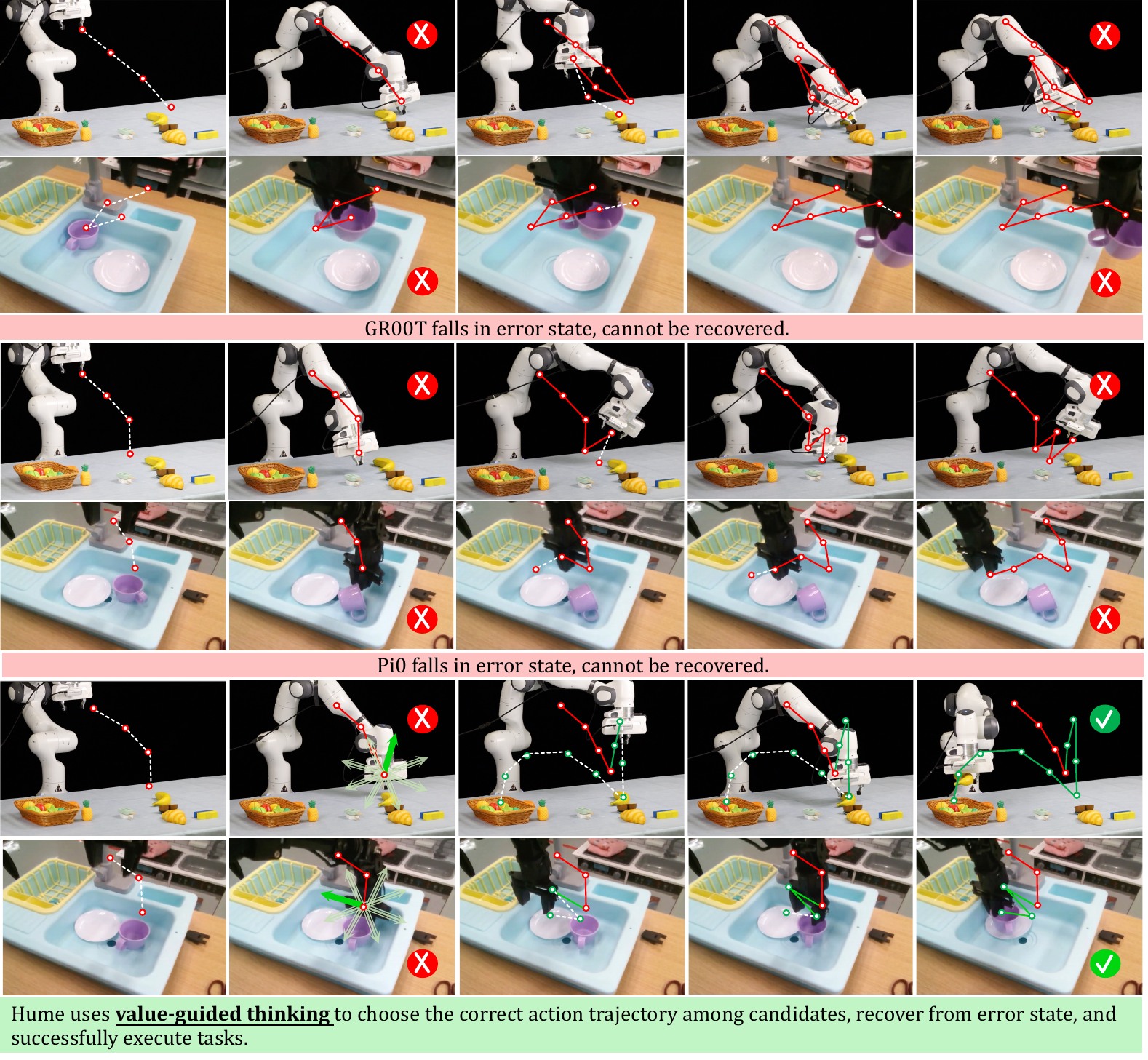}
    \end{center}
    \vspace{-1.5ex}
    \caption{\textbf{Failure Recovery of Hume.} When a failure occurs, such as missing the grasping position, other policies fall into a failure state, and Hume selects the correct action through value-guided thinking to help it recover from the failure state and successfully complete the task.}
    \label{fig:supp_recover}
    \vspace{-2ex}
\end{figure*}

\textbf{Detailed Results of Real-World Evaluation.} 
To ensure the reliability of real robot platform test results, we used standardized evaluation metrics. For simple tasks such as ``Lift red pepper'' on WidowX, we only tracked the overall task completion success rate. 
For more general tasks, such as ``Kitchen Banana'' on Franka, we tracked both partial and overall success rates, meaning the robot's success rate in grasping the banana and placing it at the designated location. 
For more complex long-horizon tasks, such as ``Pour water'' on Agibot G-1, we tracked the success rate of each subtask and the overall task success rate, including the robot's success in grasping the kettle, pouring water from the kettle into the cup, and placing the kettle on the pad.
In addition, to verify the model's generalization capability, we conducted tests under different lighting conditions, with various types of objects, different environmental layouts, and diverse language instructions.
For example, in the ``put green cup on the pink cloth'' task on the WidowX robot, the initial position of the green cup was significantly adjusted. In the ``Fold Shorts'' task on the Agibot G-1, shorts of different colors and materials were used for testing.

As shown in~\cref{fig:supp_recover}, we also observed \ours has the ability to recover from failures during evaluation. Compared to other methods, \ours can recover from failure states more often and complete the task successfully. 
Benefit from value-guided thinking, when the model falls into a failure state, such as missing the correct grasp position, \ours can select the correct action from a variety of candidate actions to recover from the failure and guide the robot to gradually recover from the failure state.
For common imitation policies such as $\pi_0$ and GR00T, when they enter an error state, since the observation of the error state does not appear in their training dataset, these models are easily trapped in the error state and cannot recover, resulting in the failure of the final task.
For Hume, although the error state observation also does not appear in its training dataset, it can include the correct action that recovers from the error state by repeatedly sampling the candidate actions that are not completely denoised, and then select the correct candidate to guide the model to recover based on the state-action value estimated by the value-query head, thereby achieving a strong ability to recover from failures.

\clearpage
\medskip
{
    \small
    \bibliographystyle{unsrt}
    \bibliography{main}

\begin{thebibliography}{10}

\bibitem{livision}
Xinghang Li, Minghuan Liu, Hanbo Zhang, Cunjun Yu, Jie Xu, Hongtao Wu, Chilam Cheang, Ya~Jing, Weinan Zhang, Huaping Liu, et~al.
\newblock Vision-language foundation models as effective robot imitators.
\newblock In {\em Proceedings of International Conference on Learning Representations (ICLR)}, 2024.

\bibitem{brohan2023rt}
Anthony Brohan, Noah Brown, Justice Carbajal, Yevgen Chebotar, Xi~Chen, Krzysztof Choromanski, Tianli Ding, Danny Driess, Avinava Dubey, Chelsea Finn, et~al.
\newblock Rt-2: Vision-language-action models transfer web knowledge to robotic control.
\newblock {\em arXiv preprint arXiv:2307.15818}, 2023.

\bibitem{liu2024fastumi}
Kehui Liu, Chuyue Guan, Zhongjie Jia, Ziniu Wu, Xin Liu, Tianyu Wang, Shuai Liang, Pengan Chen, Pingrui Zhang, Haoming Song, et~al.
\newblock Fastumi: A scalable and hardware-independent universal manipulation interface with dataset.
\newblock {\em arXiv e-prints}, pages arXiv--2409, 2024.

\bibitem{kim2024openvla}
Moo~Jin Kim, Karl Pertsch, Siddharth Karamcheti, Ted Xiao, Ashwin Balakrishna, Suraj Nair, Rafael Rafailov, Ethan Foster, Grace Lam, Pannag Sanketi, et~al.
\newblock Openvla: An open-source vision-language-action model.
\newblock {\em arXiv preprint arXiv:2406.09246}, 2024.

\bibitem{gao2025learning}
Xianqiang Gao, Pingrui Zhang, Delin Qu, Dong Wang, Zhigang Wang, Yan Ding, and Bin Zhao.
\newblock Learning 2d invariant affordance knowledge for 3d affordance grounding.
\newblock In {\em Proceedings of the AAAI Conference on Artificial Intelligence}, volume~39, pages 3095--3103, 2025.

\bibitem{pertsch2025fast}
Karl Pertsch, Kyle Stachowicz, Brian Ichter, Danny Driess, Suraj Nair, Quan Vuong, Oier Mees, Chelsea Finn, and Sergey Levine.
\newblock Fast: Efficient action tokenization for vision-language-action models.
\newblock {\em arXiv preprint arXiv:2501.09747}, 2025.

\bibitem{yao2024improving}
Yuanqi Yao, Gang Wu, Kui Jiang, Siao Liu, Jian Kuai, Xianming Liu, and Junjun Jiang.
\newblock Improving domain generalization in self-supervised monocular depth estimation via stabilized adversarial training.
\newblock In {\em European Conference on Computer Vision}, pages 183--201. Springer, 2024.

\bibitem{zhang2025moma}
Pingrui Zhang, Xianqiang Gao, Yuhan Wu, Kehui Liu, Dong Wang, Zhigang Wang, Bin Zhao, Yan Ding, and Xuelong Li.
\newblock Moma-kitchen: A 100k+ benchmark for affordance-grounded last-mile navigation in mobile manipulation.
\newblock {\em arXiv preprint arXiv:2503.11081}, 2025.

\bibitem{gao2023orla}
Kai Gao, Zhaxizhuoma, Yan Ding, Shiqi Zhang, and Jingjin Yu.
\newblock Orla*: Mobile manipulator-based object rearrangement with lazy a star.
\newblock {\em arXiv preprint arXiv:2309.13707}, 2023.

\bibitem{zhang2024decentralized}
Yang Zhang, Chenjia Bai, Bin Zhao, Junchi Yan, Xiu Li, and Xuelong Li.
\newblock Decentralized transformers with centralized aggregation are sample-efficient multi-agent world models.
\newblock {\em arXiv preprint arXiv:2406.15836}, 2024.

\bibitem{geminiroboticsteam2025geminiroboticsbringingai}
Gemini~Robotics Team, Saminda Abeyruwan, Joshua Ainslie, Jean-Baptiste Alayrac, Montserrat~Gonzalez Arenas, Travis Armstrong, Ashwin Balakrishna, Robert Baruch, Maria Bauza, Michiel Blokzijl, et~al.
\newblock Gemini robotics: Bringing ai into the physical world.
\newblock {\em arXiv preprint arXiv:2503.20020}, 2025.

\bibitem{black2024pi_0}
Kevin Black, Noah Brown, Danny Driess, Adnan Esmail, Michael Equi, Chelsea Finn, Niccolo Fusai, Lachy Groom, Karol Hausman, Brian Ichter, et~al.
\newblock A vision-language-action flow model for general robot control.
\newblock {\em arXiv preprint arXiv:2410.24164}, 2024.

\bibitem{kahneman2011thinking}
Daniel Kahneman.
\newblock Thinking, fast and slow.
\newblock {\em Farrar, Straus and Giroux}, 2011.

\bibitem{liu2024coherent}
Kehui Liu, Zixin Tang, Dong Wang, Zhigang Wang, Xuelong Li, and Bin Zhao.
\newblock Coherent: Collaboration of heterogeneous multi-robot system with large language models.
\newblock {\em arXiv preprint arXiv:2409.15146}, 2024.

\bibitem{xia2024kinematic}
Wenke Xia, Dong Wang, Xincheng Pang, Zhigang Wang, Bin Zhao, Di~Hu, and Xuelong Li.
\newblock Kinematic-aware prompting for generalizable articulated object manipulation with llms.
\newblock In {\em 2024 IEEE International Conference on Robotics and Automation (ICRA)}, pages 2073--2080. IEEE, 2024.

\bibitem{wang2025more}
Dewei Wang, Xinmiao Wang, Xinzhe Liu, Jiyuan Shi, Yingnan Zhao, Chenjia Bai, and Xuelong Li.
\newblock More: Mixture of residual experts for humanoid lifelike gaits learning on complex terrains.
\newblock {\em arXiv preprint arXiv:2506.08840}, 2025.

\bibitem{wang2025skillnavenhancednavigationversatile}
Dewei Wang, Chenjia Bai, Chenhui Li, Jiyuan Shi, Yan Ding, Chi Zhang, and Bin Zhao.
\newblock Skill-nav: Enhanced navigation with versatile quadrupedal locomotion via waypoint interface.
\newblock {\em arXiv preprint arXiv:2506.21853}, 2025.

\bibitem{xia2025roboticpolicylearninghumanassisted}
Wenke Xia, Yichu Yang, Hongtao Wu, Xiao Ma, Tao Kong, and Di~Hu.
\newblock Robotic policy learning via human-assisted action preference optimization.
\newblock {\em arXiv preprint arXiv:2506.07127}, 2025.

\bibitem{xia2025phoenix}
Wenke Xia, Ruoxuan Feng, Dong Wang, and Di~Hu.
\newblock Phoenix: A motion-based self-reflection framework for fine-grained robotic action correction.
\newblock In {\em Proceedings of the Computer Vision and Pattern Recognition Conference}, pages 6981--6990, 2025.

\bibitem{yao2025think}
Yuanqi Yao, Siao Liu, Haoming Song, Delin Qu, Qizhi Chen, Yan Ding, Bin Zhao, Zhigang Wang, Xuelong Li, and Dong Wang.
\newblock Think small, act big: Primitive prompt learning for lifelong robot manipulation.
\newblock {\em arXiv preprint arXiv:2504.00420}, 2025.

\bibitem{Lv_2025_CVPR}
Qi~Lv, Hao Li, Xiang Deng, Rui Shao, Yinchuan Li, Jianye Hao, Longxiang Gao, Michael~Yu Wang, and Liqiang Nie.
\newblock Spatial-temporal graph diffusion policy with kinematic modeling for bimanual robotic manipulation.
\newblock In {\em Proceedings of the Computer Vision and Pattern Recognition Conference (CVPR)}, pages 17394--17404, June 2025.

\bibitem{wei2022chain}
Jason Wei, Xuezhi Wang, Dale Schuurmans, Maarten Bosma, Fei Xia, Ed~Chi, Quoc~V Le, Denny Zhou, et~al.
\newblock Chain-of-thought prompting elicits reasoning in large language models.
\newblock In Alice~H. Oh, Alekh Agarwal, Danielle Belgrave, and Kyunghyun Cho, editors, {\em Advances in Neural Information Processing Systems}, volume~35, pages 24824--24837, 2022.

\bibitem{Zawalski24-ecot}
Michał Zawalski, William Chen, Karl Pertsch, Oier Mees, Chelsea Finn, and Sergey Levine.
\newblock Robotic control via embodied chain-of-thought reasoning.
\newblock {\em arXiv preprint arXiv:2407.08693}, 2024.

\bibitem{han2024dual}
ByungOk Han, Jaehong Kim, and Jinhyeok Jang.
\newblock A dual process vla: Efficient robotic manipulation leveraging vlm.
\newblock In {\em Conference on Robot Learning (CoRL)}, 2024.

\bibitem{zhang2024hirt}
Jianke Zhang, Yanjiang Guo, Xiaoyu Chen, Yen-Jen Wang, Yucheng Hu, Chengming Shi, and Jianyu Chen.
\newblock Hirt: Enhancing robotic control with hierarchical robot transformers.
\newblock {\em arXiv preprint arXiv:2410.05273}, 2024.

\bibitem{bu2024towards}
Qingwen Bu, Hongyang Li, Li~Chen, Jisong Cai, Jia Zeng, Heming Cui, Maoqing Yao, and Yu~Qiao.
\newblock Towards synergistic, generalized, and efficient dual-system for robotic manipulation.
\newblock {\em arXiv preprint arXiv:2410.08001}, 2024.

\bibitem{wen2025dexvla}
Junjie Wen, Yichen Zhu, Jinming Li, Zhibin Tang, Chaomin Shen, and Feifei Feng.
\newblock Dexvla: Vision-language model with plug-in diffusion expert for general robot control.
\newblock {\em arXiv preprint arXiv:2502.05855}, 2025.

\bibitem{bjorck2025gr00t}
Johan Bjorck, Fernando Casta{\~n}eda, Nikita Cherniadev, Xingye Da, Runyu Ding, Linxi Fan, Yu~Fang, Dieter Fox, Fengyuan Hu, Spencer Huang, et~al.
\newblock Gr00t n1: An open foundation model for generalist humanoid robots.
\newblock {\em arXiv preprint arXiv:2503.14734}, 2025.

\bibitem{shi2025hirobotopenendedinstruction}
Lucy~Xiaoyang Shi, Brian Ichter, Michael Equi, Liyiming Ke, Karl Pertsch, Quan Vuong, James Tanner, Anna Walling, Haohuan Wang, Niccolo Fusai, et~al.
\newblock Hi robot: Open-ended instruction following with hierarchical vision-language-action models.
\newblock {\em arXiv preprint arXiv:2502.19417}, 2025.

\bibitem{helix}
Figure.
\newblock Helix: A vision-language-action model for generalist humanoid control, 2025.

\bibitem{lipman2022flow}
Yaron Lipman, Ricky~TQ Chen, Heli Ben-Hamu, Maximilian Nickel, and Matt Le.
\newblock Flow matching for generative modeling.
\newblock {\em arXiv preprint arXiv:2210.02747}, 2022.

\bibitem{lin2025beyond}
Yexiong Lin, Yu~Yao, and Tongliang Liu.
\newblock Beyond optimal transport: Model-aligned coupling for flow matching.
\newblock {\em arXiv preprint arXiv:2505.23346}, 2025.

\bibitem{liu2023libero}
Bo~Liu, Yifeng Zhu, Chongkai Gao, Yihao Feng, Qiang Liu, Yuke Zhu, and Peter Stone.
\newblock Libero: Benchmarking knowledge transfer for lifelong robot learning.
\newblock {\em arXiv preprint arXiv:2306.03310}, 2023.

\bibitem{li24simpler}
Xuanlin Li, Kyle Hsu, Jiayuan Gu, Karl Pertsch, Oier Mees, Homer~Rich Walke, Chuyuan Fu, Ishikaa Lunawat, Isabel Sieh, Sean Kirmani, Sergey Levine, Jiajun Wu, Chelsea Finn, Hao Su, Quan Vuong, and Ted Xiao.
\newblock Evaluating real-world robot manipulation policies in simulation.
\newblock In {\em Proceedings of the Conference on Robot Learning (CoRL)}, 2024.

\bibitem{li2024cogact}
Qixiu Li, Yaobo Liang, Zeyu Wang, Lin Luo, Xi~Chen, Mozheng Liao, Fangyun Wei, Yu~Deng, Sicheng Xu, Yizhong Zhang, et~al.
\newblock Cogact: A foundational vision-language-action model for synergizing cognition and action in robotic manipulation.
\newblock {\em arXiv preprint arXiv:2411.19650}, 2024.

\bibitem{zeng2025FSDrive}
Shuang Zeng, Xinyuan Chang, Mengwei Xie, Xinran Liu, Yifan Bai, Zheng Pan, Mu~Xu, and Xing Wei.
\newblock Futuresightdrive: Thinking visually with spatio-temporal cot for autonomous driving.
\newblock {\em arXiv preprint arXiv:2505.17685}, 2025.

\bibitem{zheng2024tracevla}
Ruijie Zheng, Yongyuan Liang, Shuaiyi Huang, Jianfeng Gao, Hal Daum{\'e}~III, Andrey Kolobov, Furong Huang, and Jianwei Yang.
\newblock Tracevla: Visual trace prompting enhances spatial-temporal awareness for generalist robotic policies.
\newblock {\em arXiv preprint arXiv:2412.10345}, 2024.

\bibitem{pmlr-v235-lv24a}
Qi~Lv, Hao Li, Xiang Deng, Rui Shao, Michael~Y Wang, and Liqiang Nie.
\newblock {R}obo{MP}$^2$: A robotic multimodal perception-planning framework with multimodal large language models.
\newblock In Ruslan Salakhutdinov, Zico Kolter, Katherine Heller, Adrian Weller, Nuria Oliver, Jonathan Scarlett, and Felix Berkenkamp, editors, {\em Proceedings of the 41st International Conference on Machine Learning}, volume 235 of {\em Proceedings of Machine Learning Research}, pages 33558--33574. PMLR, 21--27 Jul 2024.

\bibitem{li2024towards}
Xinghang Li, Peiyan Li, Minghuan Liu, Dong Wang, Jirong Liu, Bingyi Kang, Xiao Ma, Tao Kong, Hanbo Zhang, and Huaping Liu.
\newblock Towards generalist robot policies: What matters in building vision-language-action models.
\newblock {\em arXiv preprint arXiv:2412.14058}, 2024.

\bibitem{zheng2025universal}
Jinliang Zheng, Jianxiong Li, Dongxiu Liu, Yinan Zheng, Zhihao Wang, Zhonghong Ou, Yu~Liu, Jingjing Liu, Ya-Qin Zhang, and Xianyuan Zhan.
\newblock Universal actions for enhanced embodied foundation models.
\newblock {\em arXiv preprint arXiv:2501.10105}, 2025.

\bibitem{liu2023learning}
Yang Liu, Zhaoyang Xia, Mengyang Zhao, Donglai Wei, Yuzheng Wang, Siao Liu, Bobo Ju, Gaoyun Fang, Jing Liu, and Liang Song.
\newblock Learning causality-inspired representation consistency for video anomaly detection.
\newblock In {\em Proceedings of the 31st ACM international conference on multimedia}, pages 203--212, 2023.

\bibitem{chen2023pali}
Xi~Chen, Josip Djolonga, Piotr Padlewski, Basil Mustafa, Soravit Changpinyo, Jialin Wu, Carlos~Riquelme Ruiz, Sebastian Goodman, Xiao Wang, Yi~Tay, et~al.
\newblock Pali-x: On scaling up a multilingual vision and language model.
\newblock In {\em Proceedings of the IEEE/CVF Conference on Computer Vision and Pattern Recognition (CVPR)}, 2024.

\bibitem{karamcheti2024prismatic}
Siddharth Karamcheti, Suraj Nair, Ashwin Balakrishna, Percy Liang, Thomas Kollar, and Dorsa Sadigh.
\newblock Prismatic vlms: Investigating the design space of visually-conditioned language models.
\newblock In {\em Proceedings of the International Conference on Machine Learning (ICML)}, 2024.

\bibitem{o2024open}
Open X-Embodiment Collaboration, Abby O’Neill, Abdul Rehman, Abhiram Maddukuri, Abhishek Gupta, Abhishek Padalkar, Abraham Lee, Acorn Pooley, Agrim Gupta, Ajay Mandlekar, Ajinkya Jain, et~al.
\newblock Open x-embodiment: Robotic learning datasets and rt-x models.
\newblock In {\em Proceedings of the IEEE International Conference on Robotics and Automation (ICRA)}, 2024.

\bibitem{shinn2023reflexionlanguageagentsverbal}
Noah Shinn, Federico Cassano, Ashwin Gopinath, Karthik Narasimhan, and Shunyu Yao.
\newblock Reflexion: Language agents with verbal reinforcement learning.
\newblock {\em Advances in Neural Information Processing Systems}, 36:8634--8652, 2023.

\bibitem{hao2023reasoninglanguagemodelplanning}
Shibo Hao, Yi~Gu, Haodi Ma, Joshua~Jiahua Hong, Zhen Wang, Daisy~Zhe Wang, and Zhiting Hu.
\newblock Reasoning with language model is planning with world model.
\newblock {\em arXiv preprint arXiv:2305.14992}, 2023.

\bibitem{shao2024deepseekmathpushinglimitsmathematical}
Zhihong Shao, Peiyi Wang, Qihao Zhu, Runxin Xu, Junxiao Song, Xiao Bi, Haowei Zhang, Mingchuan Zhang, YK~Li, Y~Wu, et~al.
\newblock Deepseekmath: Pushing the limits of mathematical reasoning in open language models.
\newblock {\em arXiv preprint arXiv:2402.03300}, 2024.

\bibitem{zhang2025cross}
Pingrui Zhang, Yifei Su, Pengyuan Wu, Dong An, Li~Zhang, Zhigang Wang, Dong Wang, Yan Ding, Bin Zhao, and Xuelong Li.
\newblock Cross from left to right brain: Adaptive text dreamer for vision-and-language navigation.
\newblock {\em arXiv preprint arXiv:2505.20897}, 2025.

\bibitem{NEURIPS2024_0d9dcd4e}
Tianle Gu, Zeyang Zhou, Kexin Huang, Dandan Liang, Yixu Wang, Haiquan Zhao, Yuanqi Yao, Xingge Qiao, Keqing Wang, Yujiu Yang, Yan Teng, Yu~Qiao, and Yingchun Wang.
\newblock Mllmguard: A multi-dimensional safety evaluation suite for multimodal large language models.
\newblock In A.~Globerson, L.~Mackey, D.~Belgrave, A.~Fan, U.~Paquet, J.~Tomczak, and C.~Zhang, editors, {\em Advances in Neural Information Processing Systems}, volume~37, pages 7256--7295. Curran Associates, Inc., 2024.

\bibitem{miao2023selfcheck}
Ning Miao, Yee~Whye Teh, and Tom Rainforth.
\newblock Selfcheck: Using llms to zero-shot check their own step-by-step reasoning.
\newblock {\em arXiv preprint arXiv:2308.00436}, 2023.

\bibitem{wang2025morphmark}
Zongqi Wang, Tianle Gu, Baoyuan Wu, and Yujiu Yang.
\newblock Morphmark: Flexible adaptive watermarking for large language models.
\newblock {\em arXiv preprint arXiv:2505.11541}, 2025.

\bibitem{Tree_of_Thought}
Shunyu Yao, Dian Yu, Jeffrey Zhao, Izhak Shafran, Tom Griffiths, Yuan Cao, and Karthik Narasimhan.
\newblock Tree of thoughts: Deliberate problem solving with large language models.
\newblock {\em Advances in neural information processing systems}, 36:11809--11822, 2023.

\bibitem{zhaxizhuoma2024alignbot}
Zhaxizhuoma Zhaxizhuoma, Pengan Chen, Ziniu Wu, Jiawei Sun, Dong Wang, Peng Zhou, Nieqing Cao, Yan Ding, Bin Zhao, and Xuelong Li.
\newblock Alignbot: Aligning vlm-powered customized task planning with user reminders through fine-tuning for household robots.
\newblock {\em arXiv preprint arXiv:2409.11905}, 2024.

\bibitem{chen2025setsleveragingselfverificationselfcorrection}
Jiefeng Chen, Jie Ren, Xinyun Chen, Chengrun Yang, Ruoxi Sun, and Sercan~{\"O} Ar{\i}k.
\newblock Sets: Leveraging self-verification and self-correction for improved test-time scaling.
\newblock {\em arXiv preprint arXiv:2501.19306}, 2025.

\bibitem{gao2024interpretablecontrastivemontecarlo}
Zitian Gao, Boye Niu, Xuzheng He, Haotian Xu, Hongzhang Liu, Aiwei Liu, Xuming Hu, and Lijie Wen.
\newblock Interpretable contrastive monte carlo tree search reasoning.
\newblock {\em arXiv preprint arXiv:2410.01707}, 2024.

\bibitem{luo2025o1prunerlengthharmonizingfinetuningo1like}
Haotian Luo, Li~Shen, Haiying He, Yibo Wang, Shiwei Liu, Wei Li, Naiqiang Tan, Xiaochun Cao, and Dacheng Tao.
\newblock O1-pruner: Length-harmonizing fine-tuning for o1-like reasoning pruning.
\newblock {\em arXiv preprint arXiv:2501.12570}, 2025.

\bibitem{ho2021cascaded}
Jonathan Ho, Chitwan Saharia, William Chan, David~J Fleet, Mohammad Norouzi, and Tim Salimans.
\newblock Cascaded diffusion models for high fidelity image generation.
\newblock {\em arXiv preprint arXiv:2106.15282}, 2021.

\bibitem{zhang2025revisiting}
Yang Zhang, Xinran Li, Jianing Ye, Delin Qu, Shuang Qiu, Chongjie Zhang, Xiu Li, and Chenjia Bai.
\newblock Revisiting multi-agent world modeling from a diffusion-inspired perspective.
\newblock {\em arXiv preprint arXiv:2505.20922}, 2025.

\bibitem{gu2022fdm}
Jiatao Gu, Shuangfei Zhai, Yizhe Zhang, Miguel~Angel Bautista, and Josh Susskind.
\newblock f-dm: A multi-stage diffusion model via progressive signal transformation.
\newblock {\em arXiv preprint arXiv:2210.04955}, 2022.

\bibitem{an2024bring}
Jie An, Zhengyuan Yang, Jianfeng Wang, Linjie Li, Zicheng Liu, Lijuan Wang, and Jiebo Luo.
\newblock Bring metric functions into diffusion models.
\newblock {\em arXiv preprint arXiv:2401.02414}, 2024.

\bibitem{chen2024spectral}
Bowen Chen, Liqin Liu, Chenyang Liu, Zhengxia Zou, and Zhenwei Shi.
\newblock Spectral-cascaded diffusion model for remote sensing image spectral super-resolution.
\newblock {\em IEEE Transactions on Geoscience and Remote Sensing}, 2024.

\bibitem{hur2025high}
Junhwa Hur, Charles Herrmann, Saurabh Saxena, Janne Kontkanen, Wei-Sheng Lai, Yichang Shih, Michael Rubinstein, David~J Fleet, and Deqing Sun.
\newblock High-resolution frame interpolation with patch-based cascaded diffusion.
\newblock In {\em Proceedings of the AAAI Conference on Artificial Intelligence}, volume~39, pages 3868--3876, 2025.

\bibitem{li2025cascaded}
Guangyuan Li, Yongkang Wang, Junsheng Luan, Lei Zhao, Wei Xing, Huaizhong Lin, and Binkai Ou.
\newblock Cascaded diffusion models for virtual try-on: Improving control and resolution.
\newblock In {\em Proceedings of the AAAI Conference on Artificial Intelligence}, volume~39, pages 4689--4697, 2025.

\bibitem{levine2020offlinereinforcementlearningtutorial}
Sergey Levine, Aviral Kumar, George Tucker, and Justin Fu.
\newblock Offline reinforcement learning: Tutorial, review, and perspectives on open problems.
\newblock {\em arXiv preprint arXiv:2005.01643}, 2020.

\bibitem{ptr}
Aviral Kumar, Anikait Singh, Frederik Ebert, Mitsuhiko Nakamoto, Yanlai Yang, Chelsea Finn, and Sergey Levine.
\newblock Pre-training for robots: Offline rl enables learning new tasks from a handful of trials.
\newblock In {\em Proceedings of Robotics: Science and Systems}, Daegu, Republic of Korea, 2023.

\bibitem{nakamoto2023calql}
Mitsuhiko Nakamoto, Simon Zhai, Anikait Singh, Max Sobol~Mark, Yi~Ma, Chelsea Finn, Aviral Kumar, and Sergey Levine.
\newblock Cal-ql: Calibrated offline rl pre-training for efficient online fine-tuning.
\newblock {\em Advances in Neural Information Processing Systems}, 36:62244--62269, 2023.

\bibitem{brohan2022rt}
Anthony Brohan, Noah Brown, Justice Carbajal, Yevgen Chebotar, Joseph Dabis, Chelsea Finn, Keerthana Gopalakrishnan, Karol Hausman, Alex Herzog, Jasmine Hsu, et~al.
\newblock Rt-1: Robotics transformer for real-world control at scale.
\newblock {\em arXiv preprint arXiv:2212.06817}, 2022.

\bibitem{team2024octo}
{Octo Model Team}, Dibya Ghosh, Homer Walke, Karl Pertsch, Kevin Black, Oier Mees, Sudeep Dasari, Joey Hejna, Charles Xu, Jianlan Luo, Tobias Kreiman, {You Liang} Tan, Lawrence~Yunliang Chen, Pannag Sanketi, Quan Vuong, Ted Xiao, Dorsa Sadigh, Chelsea Finn, and Sergey Levine.
\newblock Octo: An open-source generalist robot policy.
\newblock In {\em Proceedings of Robotics: Science and Systems (RSS)}, 2024.

\bibitem{wang2024scaling}
Lirui Wang, Xinlei Chen, Jialiang Zhao, and Kaiming He.
\newblock Scaling proprioceptive-visual learning with heterogeneous pre-trained transformers.
\newblock In {\em Proceedings of the Conference on Neural Information Processing System (NeurIPS)}, 2024.

\bibitem{li2023generalist}
Xinghang Li, Peiyan Li, Minghuan Liu, Dong Wang, Jirong Liu, Bingyi Kang, Xiao Ma, Tao Kong, Hanbo Zhang, and Huaping Liu.
\newblock Towards generalist robot policies: What matters in building vision-language-action models.
\newblock {\em arXiv preprint arXiv:2412.14058}, 2024.

\bibitem{qu2025spatialvlaexploringspatialrepresentations}
Delin Qu, Haoming Song, Qizhi Chen, Yuanqi Yao, Xinyi Ye, Yan Ding, Zhigang Wang, JiaYuan Gu, Bin Zhao, Dong Wang, et~al.
\newblock Spatialvla: Exploring spatial representations for visual-language-action model.
\newblock {\em arXiv preprint arXiv:2501.15830}, 2025.

\bibitem{chi2024diffusionpolicy}
Cheng Chi, Zhenjia Xu, Siyuan Feng, Eric Cousineau, Yilun Du, Benjamin Burchfiel, Russ Tedrake, and Shuran Song.
\newblock Diffusion policy: Visuomotor policy learning via action diffusion.
\newblock In {\em Proceedings of Robotics: Science and Systems (RSS)}, 2023.

\bibitem{kim2025fine}
Moo~Jin Kim, Chelsea Finn, and Percy Liang.
\newblock Fine-tuning vision-language-action models: Optimizing speed and success.
\newblock {\em arXiv preprint arXiv:2502.19645}, 2025.

\bibitem{liu2023improving}
Yixin Liu, Avi Singh, C~Daniel Freeman, John~D Co-Reyes, and Peter~J Liu.
\newblock {Improving large language model fine-tuning for solving math problems}.
\newblock {\em arXiv preprint arXiv:2310.10047}, 2023.

\bibitem{lillicrap2015continuous}
Timothy~P Lillicrap, Jonathan~J Hunt, Alexander Pritzel, Nicolas Heess, Tom Erez, Yuval Tassa, David Silver, and Daan Wierstra.
\newblock Continuous control with deep reinforcement learning.
\newblock {\em arXiv preprint arXiv:1509.02971}, 2015.

\bibitem{fujimoto2018addressing}
Scott Fujimoto, Herke van Hoof, and David Meger.
\newblock Addressing function approximation error in actor-critic methods.
\newblock In {\em International Conference on Machine Learning (ICML)}, pages 1587--1596, 2018.

\bibitem{haarnoja2018sacapps}
Tuomas Haarnoja, Aurick Zhou, Kristian Hartikainen, George Tucker, Sehoon Ha, Jie Tan, Vikash Kumar, Henry Zhu, Abhishek Gupta, Pieter Abbeel, et~al.
\newblock Soft actor-critic algorithms and applications.
\newblock {\em arXiv preprint arXiv:1812.05905}, 2018.

\end{thebibliography}
}

\end{document}